\documentclass{article}

    \PassOptionsToPackage{numbers, compress}{natbib}

\usepackage{graphicx}
\usepackage{multirow}
\usepackage{subcaption}
\usepackage{amsmath}

\usepackage[final]{neurips_data_2024}

\usepackage[utf8]{inputenc} 
\usepackage[T1]{fontenc}    
\usepackage{hyperref}       
\usepackage{url}            
\usepackage{booktabs}       
\usepackage{amsfonts}       
\usepackage{nicefrac}       
\usepackage{microtype}      
\usepackage{xcolor}         

\definecolor{darkgreen}{RGB}{0,100,0}
\definecolor{darkred}{RGB}{139,0,0}

\title{Implicit-Zoo: A Large-Scale Dataset of Neural Implicit Functions for 2D Images and 3D Scenes}

\author{ Qi Ma\textsuperscript{1,2}\space\space\space\space Danda Pani Paudel\textsuperscript{2}\space\space\space\space Ender Konukoglu\textsuperscript{1}\space\space\space\space Luc Van Gool\textsuperscript{1,2}\\
\textsuperscript{1}Computer Vision Lab, ETH Zurich\space\space\space\space \textsuperscript{2}INSAIT, Sofia University  }

\begin{document}

\maketitle

\begin{abstract}
    Neural implicit functions have demonstrated significant importance in various areas such as computer vision, graphics. Their advantages include the ability to represent complex shapes and scenes with high fidelity, smooth interpolation capabilities, and continuous representations. Despite these benefits, the development and analysis of implicit functions have been limited by the lack of comprehensive datasets and the substantial computational resources required for their implementation and evaluation. To address these challenges, we introduce "Implicit-Zoo": a large-scale dataset requiring thousands of GPU training days designed to facilitate research and development in this field. Our dataset includes diverse 2D and 3D scenes, such as CIFAR-10, ImageNet-1K, and Cityscapes for 2D image tasks, and the OmniObject3D dataset for 3D vision tasks. We ensure high quality through strict checks, refining or filtering out low-quality data. Using Implicit-Zoo, we showcase two immediate benefits as it enables to: (1) \emph{learn token locations} for transformer models; (2) \emph{directly regress} 3D cameras poses of 2D images with respect to NeRF models. This in turn leads to an \emph{improved performance} in all three task of  image classification, semantic segmentation, and 3D pose regression -- 
    thereby unlocking new avenues for research.
    Our data and implementation are available from: \href{https://github.com/qimaqi/Implicit-Zoo/}{https://github.com/qimaqi/Implicit-Zoo/}

\end{abstract}

\begin{figure}[ht]
\centering
        \centering
        \includegraphics[width=0.8\linewidth, trim={0  0  0  0},clip ]{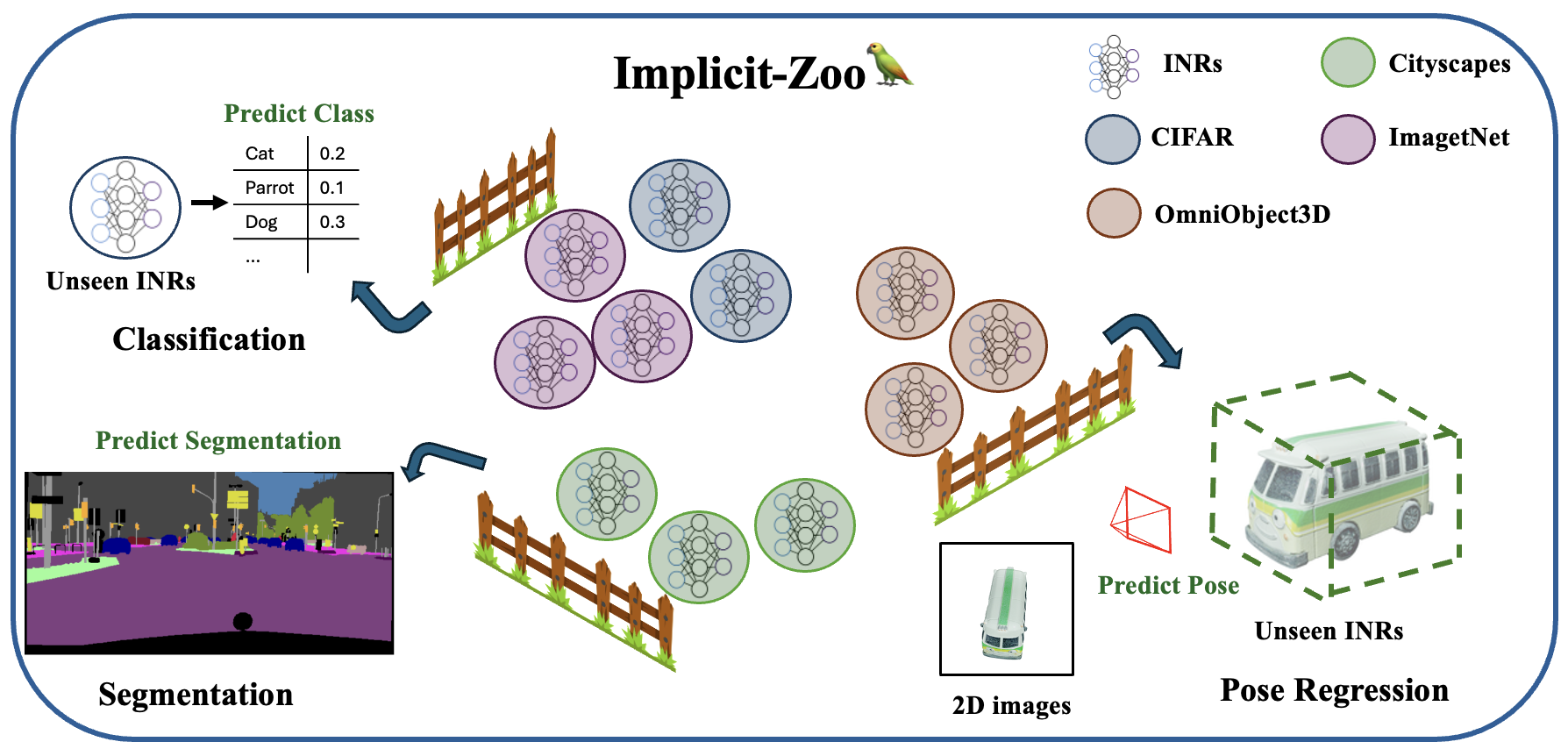}
\caption{\textbf{\textit{The Implicit-Zoo dataset and its example utilities.}} We demonstrate three tasks using \textit{Implicit-Zoo}: classification, segmentation, and 3D pose regression. Details of the problem statement can be found in section \ref{statement}. The INRs are colorized differently to indicate their training data sources. \label{fig:application}}
\end{figure}

\begin{table}[t]
\resizebox{\linewidth}{!}{%
\begin{tabular}{l|llclll}
\toprule
Method & Task & Scenes  & Model(Depth/Width)  & GPU (days) & Overall Size (GB) & PSNR \\
\midrule 
CIFAR-10 \cite{krizhevsky2009learning} & 2D & 60000  & 3 / 64  & 5.96 &  1.44  & 31.01  \\
ImageNet-1K \cite{deng2009imagenet} & 2D & 1431167  & 4 / 256  & 831.53 & 749.93 & 30.12  \\
CityScapes \cite{cordts2016cityscapes}& 2D & 23473  & 5 / 256  & 50.15 & 18.40  & 34.10  \\
Omniobject-3D \cite{wu2023omniobject3d} & 3D & 5914  & 4 / 128 & 69.81  & 5.96   & 31.54  \\
\bottomrule
\end{tabular}}
\vspace{1mm}
\caption{\textbf{Dataset Summary.} An overview of the generated datasets with their cost and quality. For the 2D task, we employ SIREN \cite{sitzmann2020implicit}, while for the 3D task, we utilize NERF \cite{mildenhall2021nerf}. Computation is carried out on the ETH Euler cluster. We report PSNR (Peak Signal-to-Noise Ratio) as  quality metric. A PSNR of 30 dB corresponds to an RGB MSE of appr. 0.03. This error level is hardly noticeable to the human eye, indicating the high fidelity in our dataset. Please refer to Figure~\ref{fig:example_imgs} for examples.}\label{tab:introduce_cost_of_data}
\end{table}

\section{Introduction}
Recent advances in modeling continuous functions implicitly using Multi-Layer Perceptrons (MLPs)~\cite{xie2022neural} have sparked great interest in many applications. Implicit neural representations (INRs) involve fitting a function $f(.)$ that maps input coordinates $x$ to their corresponding value $v_x$. For instance, in image modeling~\cite{sitzmann2020implicit}, this continuous function maps 2D pixel coordinates to RGB values. This approach has been extended to challenging 3D geometry and appearance~\cite{mildenhall2021nerf, mescheder2019occupancy, chen2019learning, park2019deepsdf}. Implicit representations offer many advantages, including strong modeling effectiveness, memory efficiency, the ability of a single architecture to represent different data modalities, compatibility with arbitrary resolutions, and differentiability.

A hindrance to further advancements in implicit function research is the lack of large-scale datasets, primarily due to their computational demands. This paper aims to bridge this gap via the \textit{Implicit-Zoo} Datasets, with access to over 1.5 million implicit functions across multiple 2D and 3D tasks.

Some INRs datasets exist in both 2D and 3D formats. However, they are limited by data scale~\cite{navon2023equivariant} and application scenarios~\cite{deluigi2023deep,ramirez2023deep}. Additionally, while using modulation~\cite{dupont2022data,bauer2023spatial} to learn the common parts of given dataset to accelerate convergence, it struggles to effectively handle unseen data samples. We will also organize and update the datasets used for concurrent work. For example, \cite{irshad2024nerfmae} use Instant-NGP \cite{M_ller_2022} to train a large amount of indoor data.

We focus on generating dataset using INRs directly on modelling image signals because of their wide-ranging applications, following the existing remarkable success~\cite{dosovitskiy2021image, xie2021segformer}. We believe our dataset will have broad applicability and make a significant contribution to the community. Additionally, we chose popular 2D datasets because they have well-established benchmarks \cite{deng2009imagenet, cordts2016cityscapes, krizhevsky2009learning}, enabling better performance comparison with other state-of-the-art algorithms.

Moreover, we developed a comprehensive 3D Implicit Neural Representations (INRs) dataset using Ombiobject3D \cite{wu2023omniobject3d} and establish the first benchmark for 3D INRs pose regression. As INRs gain the status of preferred representation for 3D scenes, determining pose from given images with trained INRs becomes crucial. \cite{lin2021barf, bian2023nope, chen2023local} aim to jointly address reconstruction and image registration challenges during the training phase of Implicit Neural Representations (INRs) without relying on pose priors.\cite{maggio2022locnerf,yen2021inerf,moreau2022lens} focus on pose regression using pretrained INRs. However, these methods only achieve convergence in scenarios with coarse pose initialization or when a scene-dependent regressor is trained, limiting their applicability to new scenes. To address this, we introduce a transformer-based approach that samples the neural radiance field to extract volume feature, integrating it with 2D images for precise pose regression. In unseen scenarios, our method achieves a rotational error of 20$^o$, with nearly 80$\%$ of poses having rotational errors below 30$^o$. Further refinement is shown by minimizing a photometric error \cite{lin2021barf}.

To effectively train our large-scale dataset, we carefully selected model sizes \ref{tab:introduce_cost_of_data} appropriate for the complexity of the data and ran a sufficient number of iterations. To maintain data quality, we conducted a second training round to guarantee that all images reached a PSNR of 30.

Thanks to the differentiability of INRs and our large-scale dataset, our transformer-based methods can efficiently optimize tokenization for various tasks. Instead of using manually designed tokenization techniques like standard patchification or volumification, our approach allows the network to learn the tokenization process directly from the large-scale dataset. We propose methods that utilize learnable 

patch centers and scales, as well as a fully learnable approach at the pixel-wise or point-wise level. Our findings indicate that this learnable tokenization significantly improves performance across multiple tasks. This research uncovers a novel direction: learnable tokenization.

In summary: our key contributions are as follows: \\
$\bullet$ We create \textit{Implicit-Zoo}, a large-scale implicit function dataset developed over almost 1000 GPU days. Through iterative refinement, incl. filtering and continuous training, we ensure its high quality.
$\bullet$ We benchmark a range of tasks using this dataset, such as 2D image classification and segmentation. Additionally, we introduce a transformer-based approach for the direct pose regression for new images in the 3D neural radiance fields. For the latter, a novel baseline is also established.

$\bullet$ By integrating learnable tokenizer, we enhance the benchmark methods across multiple tasks. 

\section{Related Work}

\subsection{Implicit neural representations (INRs)}
SIREN \cite{sitzmann2020implicit} use periodic activation functions to capture high-frequency details in images. Building on this, \cite{ramasinghe2022beyond} propose using Gaussian functions, which offer greater robustness to random initialization and require fewer parameters. \cite{saragadam2023wire} introduce a continuous complex Gabor wavelet to robustly represent natural images with high accuracy. \cite{chen2021learning} focus on continuous representation for arbitrary resolution.Implicit Neural Representations (INRs) have been shown to effectively represent scenes as occupancy \cite{mescheder2019occupancy,saragadam2023wire,sitzmann2020implicit}, object shape \cite{chabra2020deep, chen2019learning, genova2019learning}, sign distance function \cite{park2019deepsdf, liu2020dist}, 3D scene appearance and dynamics \cite{mildenhall2021nerf, xie2022neural,Chen_2023_ICCV,mueller2022instant,barron2021mipnerf,ma2023deformable,park2021nerfies,li2023dynibar} and other complex signals and problems\cite{kuznetsov2021neumip,mcginnis2024singlesubject,xu2023nesvor,sun2021coil,ma2024continuous}.

\subsection{Transformer on various computer vision tasks}
Vision Transformers (ViTs) have achieved state-of-the-art (SOTA) performance in several \textbf{Image recognition} tasks\cite{khan2022transformers} and are thus selected as our primary benchmark. It rely on the self-attention mechanism 
\cite{vaswani2017attention,touvron2021training} and require large datasets for effective training. This underscores the need for large datasets in Implicit Neural Representations (INRs) format, especially for transformer-based methods. Typically, after pre-training in either a supervised \cite{dosovitskiy2021image, touvron2022deit} or self-supervised \cite{he2021masked, caron2021emerging,devlin2018bert, hatamizadeh2021swin,gidaris2018unsupervised} manner, fine-tuning is performed on smaller datasets for downstream tasks. 
\textbf{Image segmentation,} which involves dense prediction and requires a larger number of tokens, poses challenges for ViTs with fixed token number and dimension. \cite{wang2021pyramid, carion2020endtoend, liu2021swin,wu2021cvt,huang2021shuffle,yang2021focal,wang2021pyramid,wang2022pvt,guo2022segnext} addressed this issue by hierarchical reducing token numbers and increase token feature dimension through convolution backend. Such multi-stage design perform well for also recognition task.\cite{xie2021segformer} developed a lightweight transformer encoder that incorporates sequence reduction with a lightweight MLP decoder. In the field of \textbf{pose regression,} where transformer-based methods have also shown success\cite{amini2022yolopose, periyasamy2023yolopose, amini2021t6ddirect, jantos2023poet, xiao2024effloc,song2024transbonet,Sun_2023}. Many of them build upon  \cite{carion2020endtoend} which outputs a set of tuples with fixed cardinality for detection tasks.

\subsection{Tokenizer of Transformer}
Recent advancements in vision transformers have focused on improving tokenization strategies to enhance performance and efficiency. \cite{vaswani2017attention} explored the effectiveness of tokenizers through Modulation across Tokens (MoTo) and TokenProp. The T2T-ViT model by \cite{dosovitskiy2021image} introduced progressive tokenization and an efficient backbone inspired by CNNs, leading to significant performance gains on ImageNet. \cite{ramasinghe2022beyond} proposed MSViT, a dynamic mixed-scale tokenization scheme that adapts token scales based on image content, optimizing the accuracy-complexity trade-off. \cite{he2021masked} developed token labeling, a dense supervision approach that reformulates image classification into token-level recognition tasks, greatly enhancing ViT performance and generalization on downstream tasks.

\subsection{Pose regression within Neural Radiance Field}
LENS\cite{moreau2022lens} applies novel view synthesis to the robot re-localization problem, using NeRF to render synthetic datasets that improve camera pose regression. Loc-NeRF \cite{maggio2022locnerf} combines Monte Carlo localization with NeRF to perform real-time robot localization using only an RGB camera. iNeRF \cite{yen2021inerf}inverts a trained NeRF to estimate 6DoF poses through gradient descent and propose proper ray sampling mechanism. Note that all method need heavy volumetric rendering process.

\section{The Implicit-Zoo Dataset} 
We introduce the \textit{Implicit-Zoo} dataset which includes over 1.5 million INRs and took almost 1000 GPU days for training on RTX-2080. We first describe the dataset over different tasks and generation process, incl. some necessary changes and the cost for large scale training of implicit functions. Then we go though the quality check and lastly we describe the licence for the released dataset.





\subsection{Dataset generation}

\begin{figure}
\centering
        \centering
        \includegraphics[width=\linewidth]{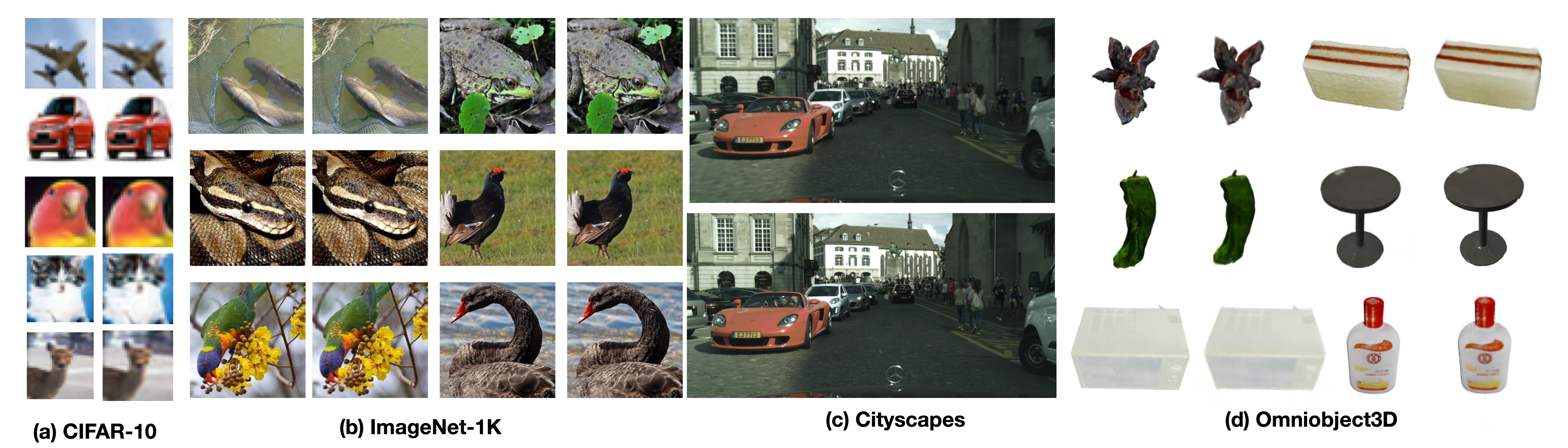}
\caption{\textbf{Examples of images from INRs.} We present visual comparisons of example image pairs from our Implicit-Zoo dataset. The original (left/top) and the reconstruction from INRs (right/bottom) images are presented in pairs, showcasing similar visual quality. Please zoom in for details.\label{fig:example_imgs}}
\end{figure}

\textbf{CIFAR-10-INRs \cite{krizhevsky2009learning}}  We test our algorithm with CIFAR-10. We employ a 3-layer SIREN MLP with a width of 64, without positional encoding \cite{sitzmann2020implicit}. For all 2D image tasks the images are normalized using a mean of $[0.485, 0.456, 0.406]$ and a standard deviation of $[0.229, 0.224, 0.225]$. We use the Adam optimizer with a learning rate of 1e-3 and a cosine learning rate scheduler with minimum learning 1e-5 without warmup. Each image is trained for 1000 iterations, requiring 8.58 secs in total. \

\textbf{ImageNet-1K-INRs} \cite{deng2009imagenet} Following the standard procedure for ImageNet-1K classification, we resize images to 256x256 and then center crop them to 224x224 for further processing. To better fit the high-res images, we employ a 4-layer 256 width SIREN. The normalization parameters are consistent with those used for CIFAR-10. We increase the training iterations to 3000, each taking 50.24 secs.

\textbf{CityScapes-INRs} \cite{cordts2016cityscapes} The Cityscapes dataset focuses on semantic understanding of urban street scenes, requiring exceptionally high image quality for pixel-wise classification. We resize the images from 1024x2048 to 320x640 pixels. To handle the details, we use a 5-layer SIREN model with a width of 256 and 3000 training iterations. The cost for training individual INR is 184.3 secs.

\textbf{Omniobject3D-INRs} \cite{wu2023omniobject3d} The Omniobject3D dataset comprises 5998 objects across 190 daily categories. For training, we utilize the official implementation of NeRF\cite{mildenhall2021nerf}, assuming a white background. Our setup includes a range of [2, 6], 64 samples per ray, and 2048 rays in one batch. Initially, we resize the images from 800x800 to 400x400 pixels and use first 96 views to generate the neural radiance field. The time cost for training a single scene is 1019 seconds.

\subsection{Dataset quality control}\label{quality}
To account for varying convergence rates across images and to optimize time efficiency, we have developed a third-phase training framework. First, we conduct basic training with a predefined number of iterations and a learning rate scheduler. Second, for data that fails to achieve a PSNR of 30 dB after basic training, we initiate extended training, which allows for up to three times the number of iterations as the basic phase. An early-stopping mechanism is incorporated into the extended training to prevent unnecessary computation.In addition, we conducted a thorough data check after, performing further training on all data that still had not achieved a PSNR of 30. For this round we train till the PSNR threshold is met.
Some examples of data is shown in Fig ~\ref{fig:example_imgs}.

\subsection{Data protection and licence}\label{licence}
Implicit-Zoo is constructed using several popular RGB datasets. \textbf{CIFAR10}: MIT-License, we are allowed to redistribute under same terms and we will release it on Kaggle. \textbf{ImageNet-1K}: Similar to CIFAR we will redistribute INRs on Kaggle, consistent with the original distribution protocols. \textbf{Cityscapes}: Non-commercial purposes licence, This dataset will be hosted on the Cityscapes team's official \href{https://www.cityscapes-dataset.com/}{website},where users must agree to the specified terms of use. \textbf{Omniobject3D}: CC BY 4.0 licence, we are permitted to redistribute it in our format directly under the same terms.

\begin{figure}[t]
\centering
        \centering
        \includegraphics[width=\linewidth, trim={0  0  0  0},clip ]{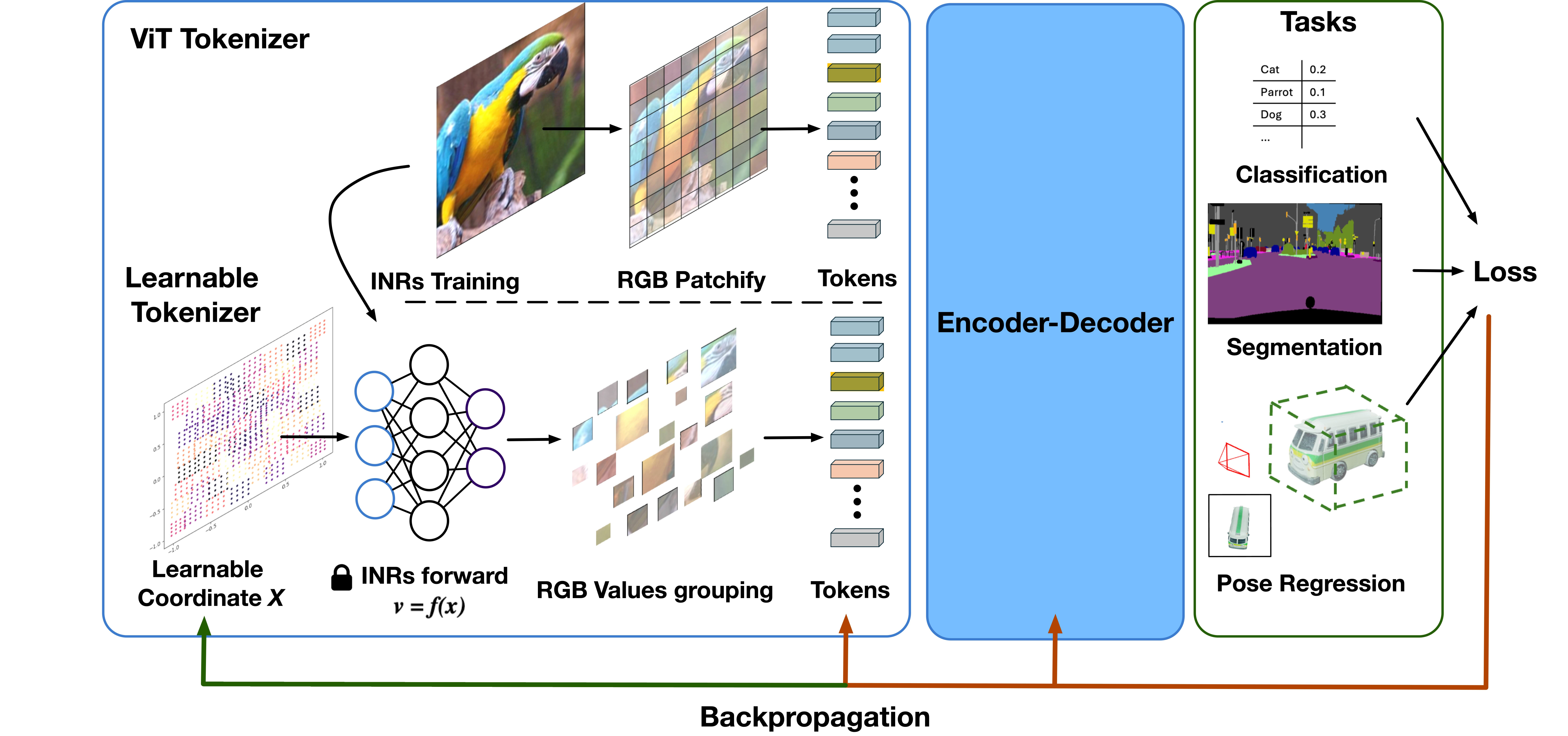}
\caption{\textbf{Illustration of learnable tokenizer.} Instead of retrieve RGB value from images we query learnable coordinates to pre-trained freezed INRs and grouping RGB values to create tokens. Note that during backpropogation the \textcolor{darkgreen}{Coordinate $x$} will also be jointly optimized with \textcolor{darkred}{ViT modules}.\label{fig:learnable_token}}. 
\end{figure}

\section{Dataset Applications}\label{statement}
\subsection{Different tasks statement}

\textbf{Classification:} We provide a dataset of INRs $\{f_i\}_{i=1}^{N}$ and associated labels {$\{y_i\}_{i=1}^{N}$}, ${y_i \in \{0,1,2...C-1\}}$. $C$ is the total classes number and $N$ is the size of dataset. The goal is to learn a model $g:f_i$ $\rightarrow$ $y_i$ that accurately predicts the label $y_i$ for a given INRs $f_i$. Recall, the RGB values at the 2d coordinates $x$ is obtained using INRs such that $v_x=f_i(x)$.

\textbf{Segmentation:} Similar to classification, in segmentation the model needs to predict densely the pixel-wise labels.
To achieve this, coordinate querying is crucial for establishing pixel-wise correspondence.

\begin{figure}[ht]
\centering
        \centering
        \includegraphics[width=\linewidth, trim={0  0  0  0},clip ]{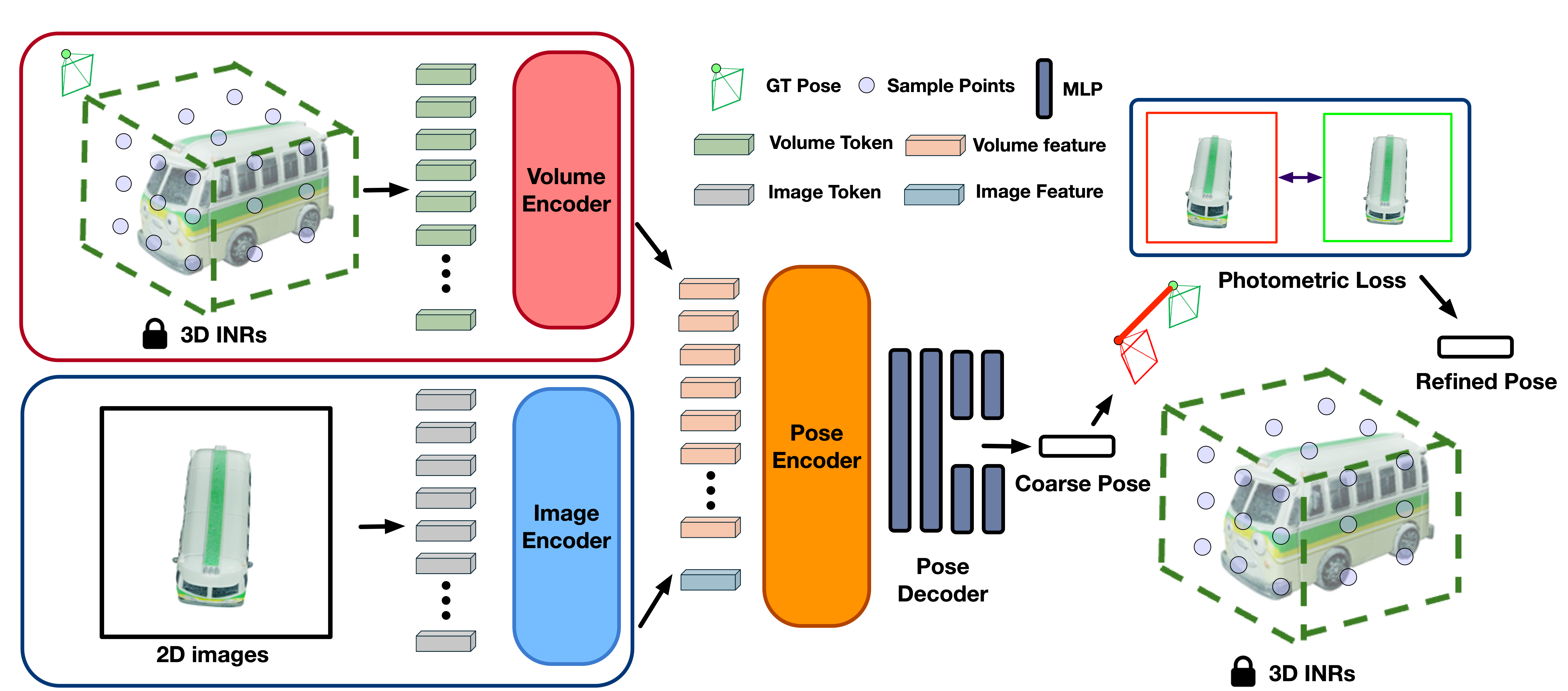}
\caption{\textbf{Illustration of proposed pose regressor.} We process 3D volume features and 2D image features with transformer-based encoder and output coarse poses. For further refinement, we freeze the 3D INRs and optimize the pose by minimizing the photometric error.
\label{fig:pose_regressor}}
\end{figure}

\textbf{Pose Regression:} In pose regression, we localize an image $I$  with respect to the 3D scene represented by its INR $f(.)$. To do so, we train a neural network $g:(f,I) \rightarrow \theta$, where $\theta$ represents the 6D pose parameters. In other words, we directly regress the pose of the image $I$ using a neural network that leverages the INRs. Thanks to the scale of our dataset, we can train such network which also generalizes beyond the training scenes. Furthermore, our dataset allows to meaningfully learn the tokens' locations.  We refer this process as learnable organization, which is presented below. 


\subsection{Learnable tokenizer} Vision Transformers treat input as sequences of patches \cite{dosovitskiy2021image,liu2021swin} or volumes \cite{feichtenhofer2022masked} -- which are also referred as tokens -- from \emph{fixed and handcrafted locations $x$}. Let a token be $t$ and $z = (x,v_x)$ be a tuple of location and the corresponding value. Then for a set $\mathcal{Z} = \{z_k\}_{k=1}^M$ of $M$ tuples, we create the token $t$ by using a learnable function $T(.)$ such that $t = T(\mathcal{Z})$. It is important to note that the \emph{INRs make the location $x$ learnable with respect to token $t$}. Thanks to the differentiability and scalability of our Dataset, we propose to jointly optimize tokenizer by making $x$ learnable with other ViT modules as shown in Fig~\ref{fig:pose_regressor}. In our experiments, we use convolution operation $T(.)$ as tokenizer. To make this approach effective is challenging. It requires specific \emph{RGB grouping methods} for both 2D and 3D INRs, \emph{ensuring differentiability} at every step especially for data augmentation. Please refer to the supplementary for details of our differentiable augmentation scheme.

\subsubsection{RGB grouping strategies }\label{rgb_grouping}
\begin{figure}[h]
    \centering
    \begin{subfigure}[t]{0.19\textwidth}
        \includegraphics[width=\textwidth]{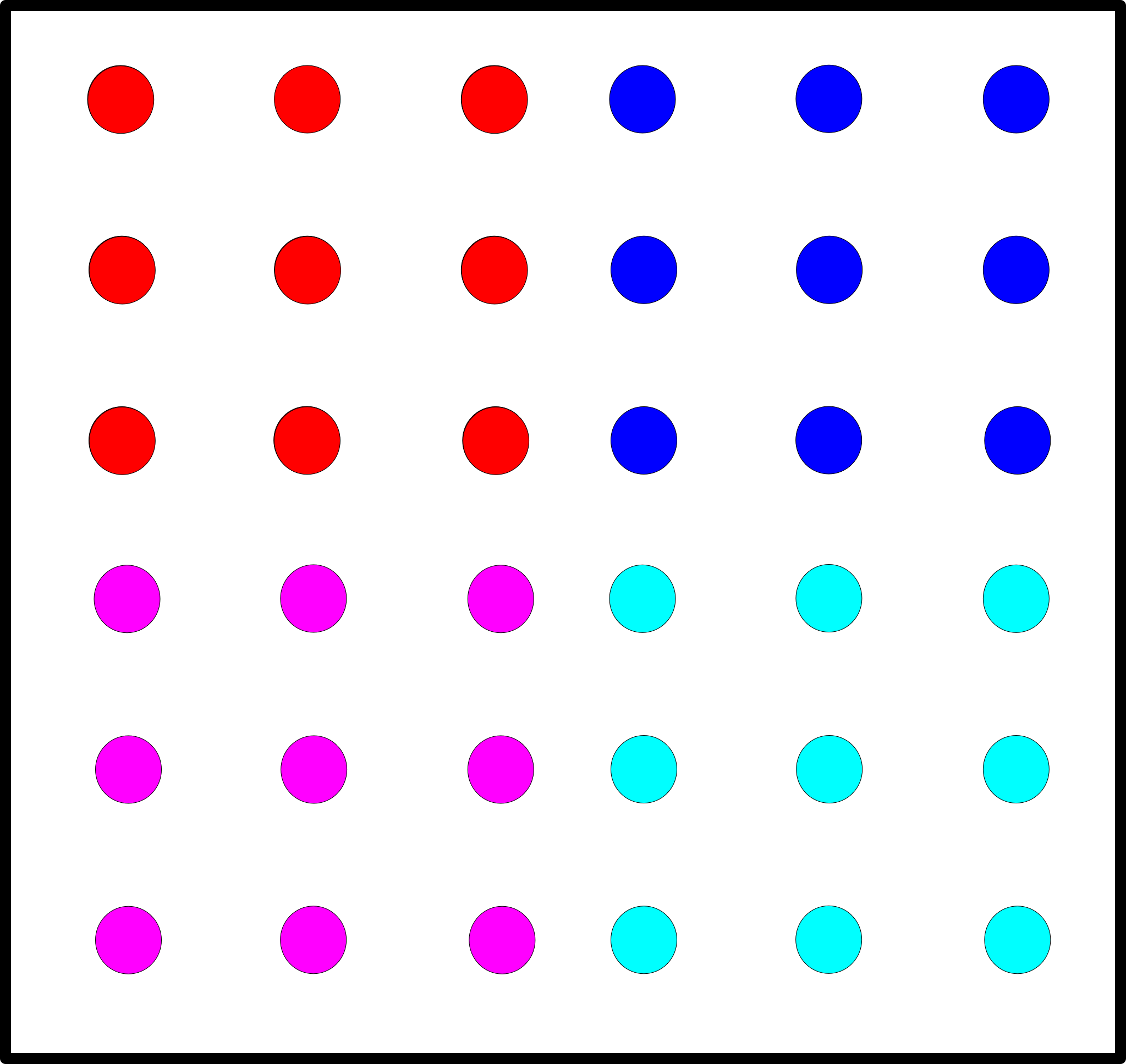}
        \caption{Uniform Patchify}
        \label{fig:group_vit}
    \end{subfigure}
    \hfill
    \begin{subfigure}[t]{0.19\textwidth}
        \includegraphics[width=\textwidth]{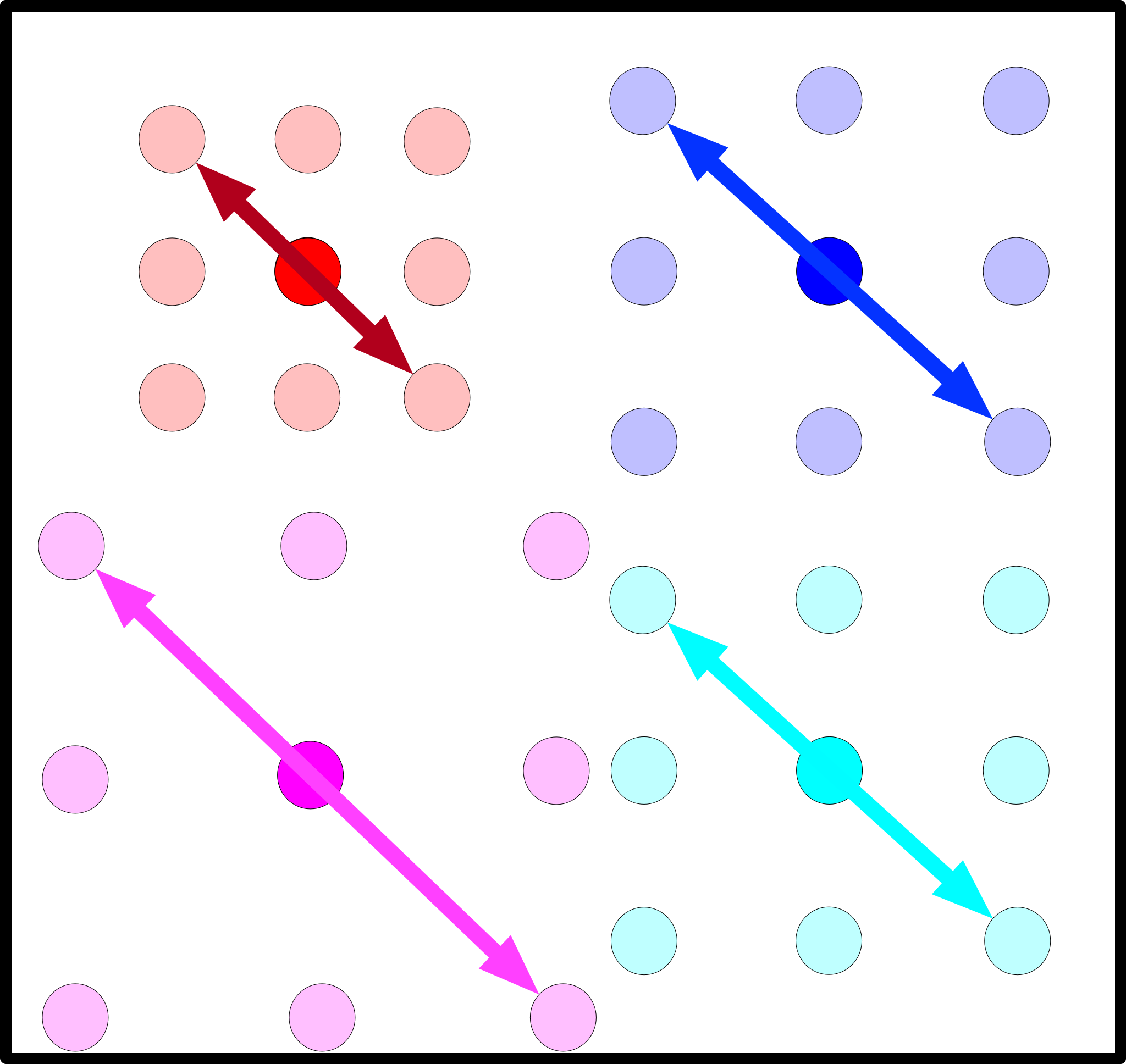}
        \caption{Uniform Centers + Learnable Scaling}
        \label{fig:group_vit_scale}
    \end{subfigure}
    \hfill
    \begin{subfigure}[t]{0.19\textwidth}
        \includegraphics[width=\textwidth]{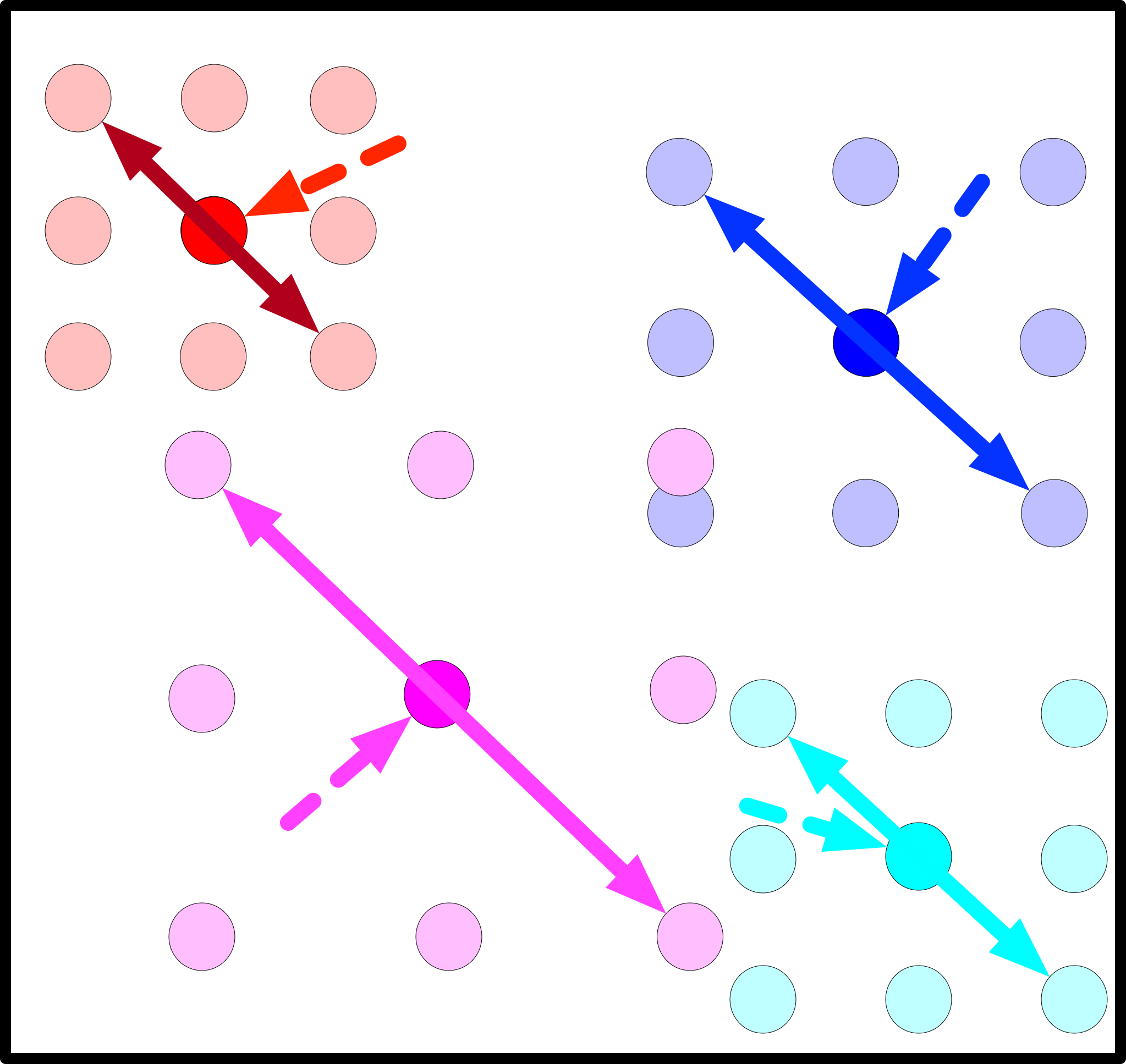}
        \caption{Learnable Centers + Learnable Scaling}
        \label{fig:group_vit_scale_move}
    \end{subfigure}
    \hfill
    \begin{subfigure}[t]{0.19\textwidth}
        \includegraphics[width=\textwidth]{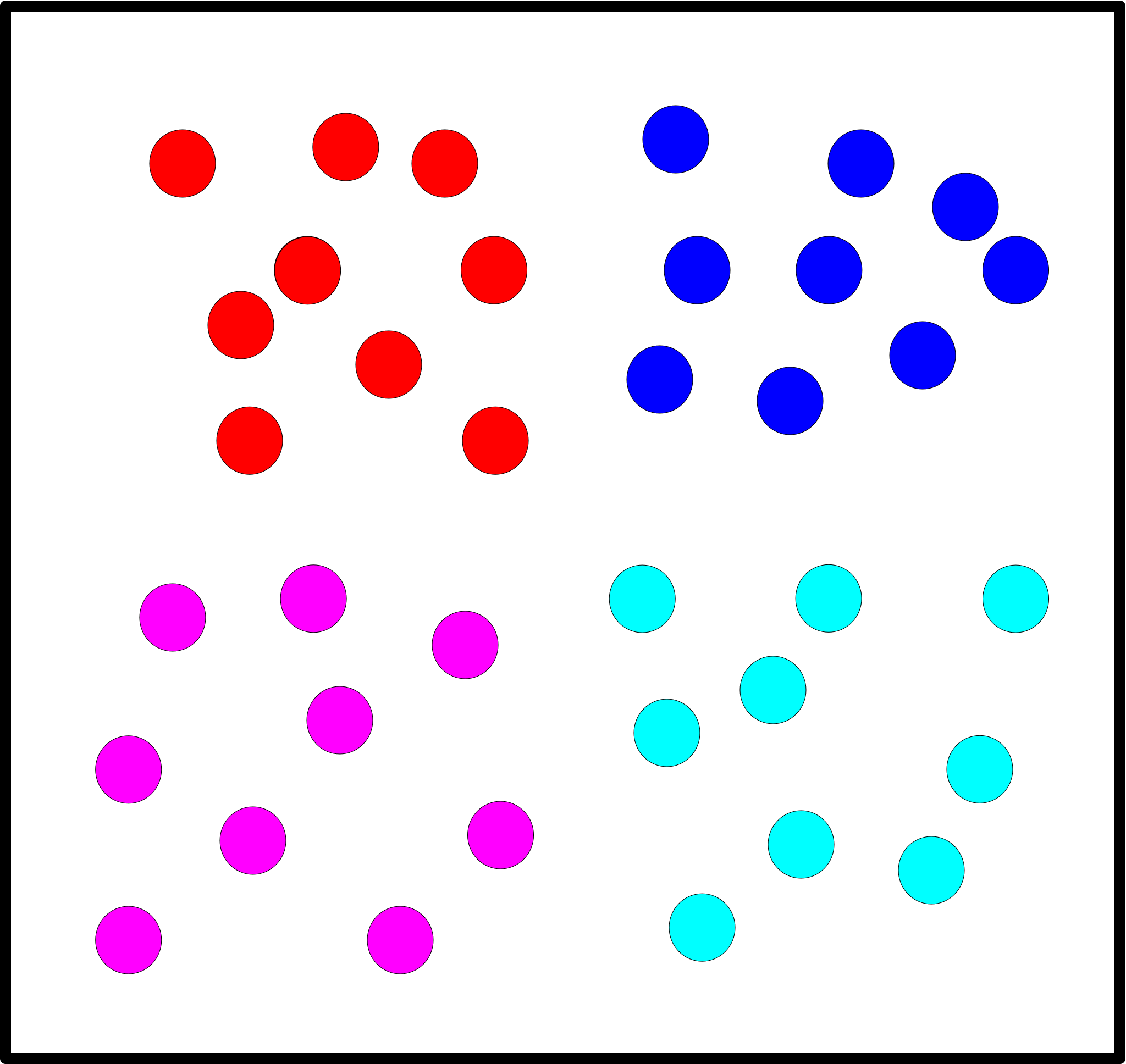}
        \caption{Learnable Pixels}
        \label{fig:group_vit_free}
    \end{subfigure}
    \hfill
    \begin{subfigure}[t]{0.19\textwidth}
        \includegraphics[width=\textwidth]{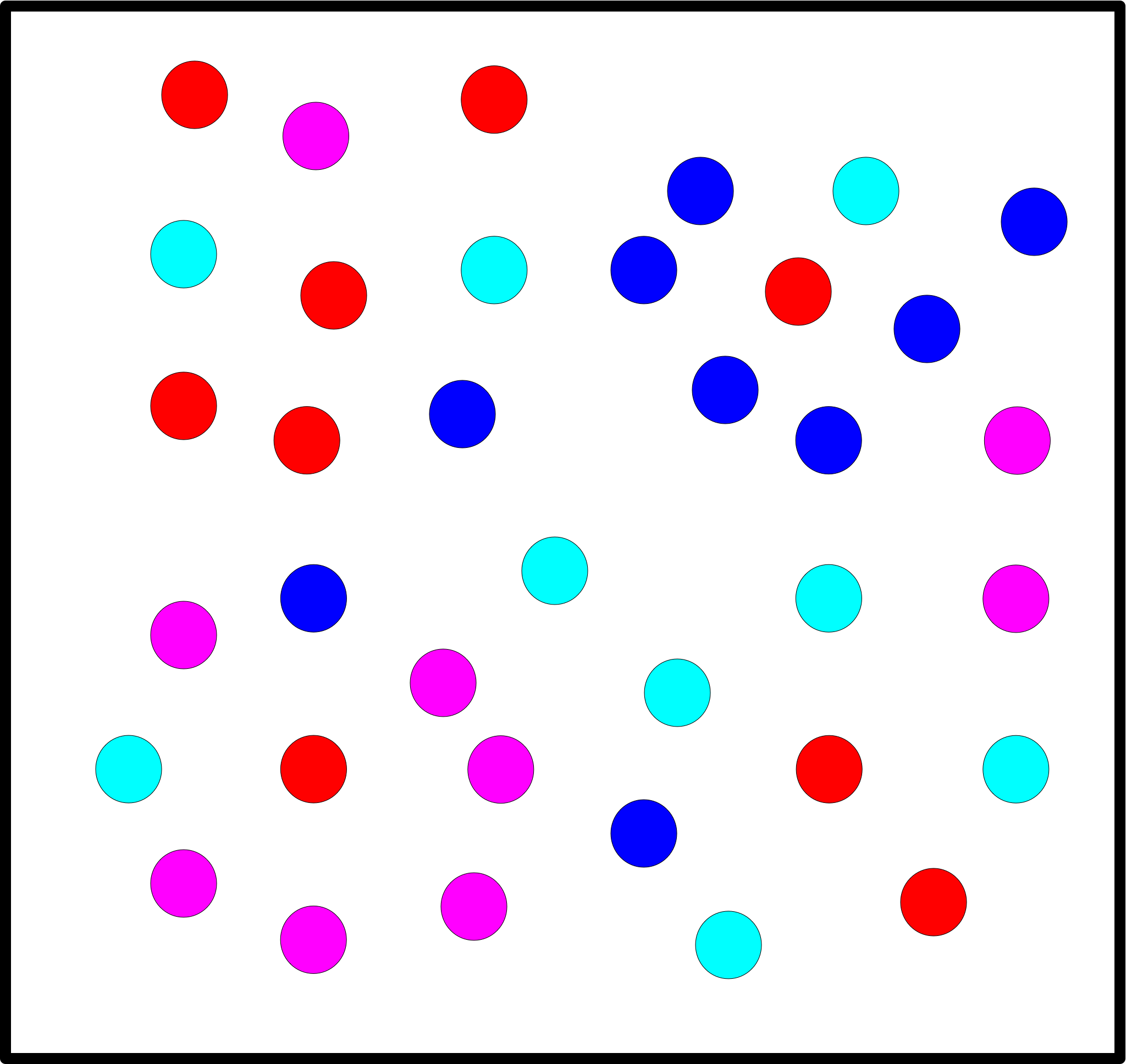}
        \caption{Random Initialize  +  Learnable Pixels}
        \label{fig:group_random}
    \end{subfigure}
    \hfill
    \caption{\textbf{Different RGB grouping strategies.} We visualize the proposed RGB grouping strategies with patch size 3. Coordinates with same color will be grouped into the same token. We abbreviate  these approaches as (b) ``S", (c) ``LC", (d) ``LP",  and (e)  ``LP+rand".}
    \label{fig:grouping}
    
\end{figure}

\begin{figure}[h]
    \centering
    \begin{subfigure}[t]{0.24\textwidth}
        \includegraphics[width=\textwidth,trim={50 50 50 50},clip]{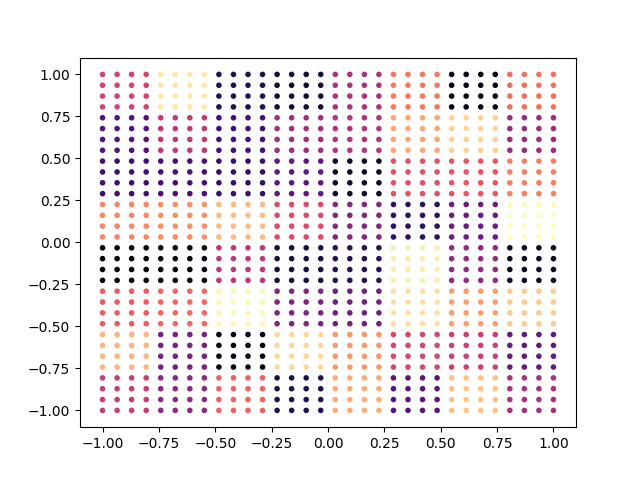}
        \caption{Uniform Patchify}
        \label{fig:vis_group_vit}
    \end{subfigure}
    \hfill
    \begin{subfigure}[t]{0.24\textwidth}
        \includegraphics[width=\textwidth,trim={50 50 50 50},clip]{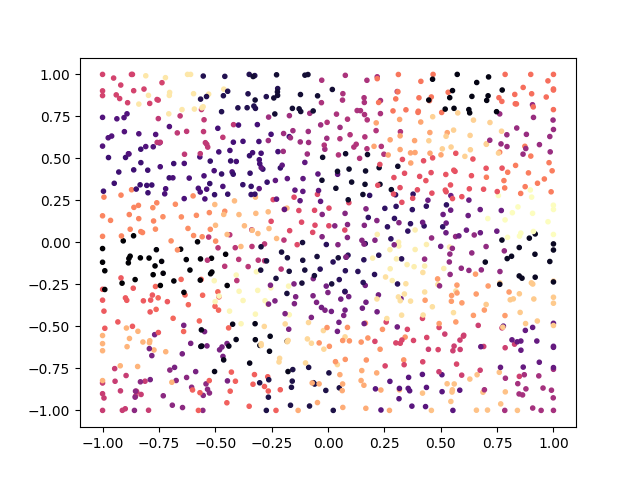}
        \caption{Uniform Centers  \protect \\  + Learnable Scaling}
        \label{fig:vis_group_vit_free}
    \end{subfigure}
    \hfill
    \begin{subfigure}[t]{0.24\textwidth}
        \includegraphics[width=\textwidth,trim={50 50 50 50},clip]{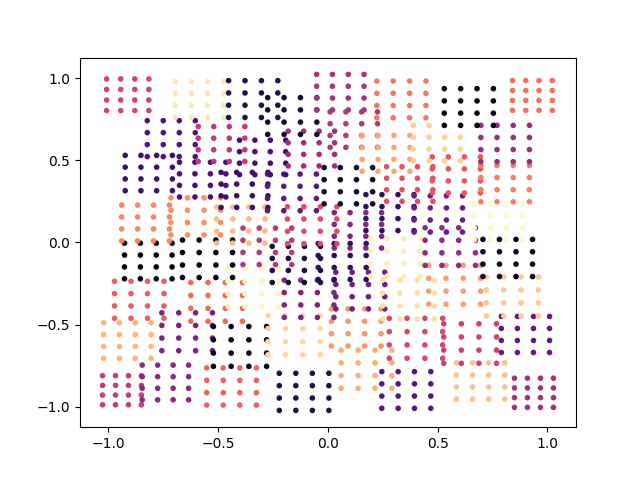}
        \caption{Learnable Centers  \protect \\ + Learnable Scaling}
        \label{fig:vis_moving_center_patch}
    \end{subfigure}
    \hfill
    \begin{subfigure}[t]{0.24\textwidth}
        \includegraphics[width=\textwidth,trim={50 50 50 50},clip]{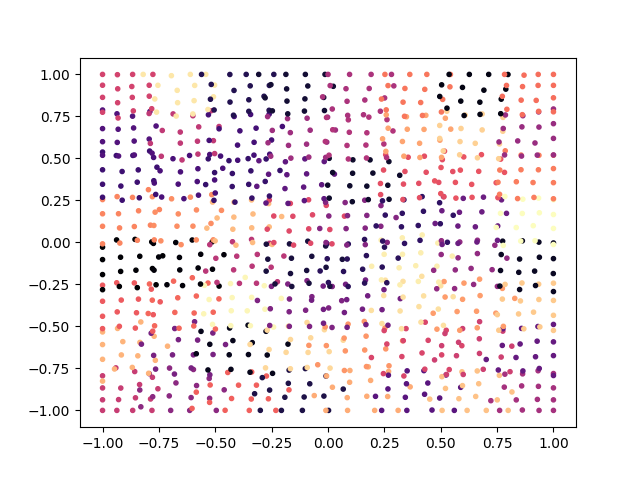}
        \caption{Learnable Pixels}
        \label{fig:vis_reg}
    \end{subfigure}
    \hfill
    \caption{\textbf{Learned tokens in CIFAR-10 INRs settings.} Different strategies corresponding to Figure~\ref{fig:grouping}. Same color (chosen randomly for visualization) indicates grouping into the same token.}
    \label{fig:vis_grouping}
\end{figure}

ViT use uniform patches of the same size, as shown in Fig. \ref{fig:group_vit}. In contrast, we propose learnable scaling and learnable center location to handle multi-resolution and allow patches to more important area, shown in Fig \ref{fig:group_vit_scale},\ref{fig:group_vit_scale_move}. Furthermore we propose all pixels coordinate are learnable \ref{fig:group_vit_free}. All above mentioned coordinates are initialized with the original RGB pixel coordinates. We also investigate the random initialized shown in Fig and \ref{fig:group_random}. Corresponding quantitative and qualitative results please refer to 
\ref{tab:cifar_main_results} and \ref{fig:vis_grouping}.

\begin{table}[h]
\centering
\centering
\begin{tabular}{l|ccc}
\toprule
Method & Acc $\uparrow$ & Precision $\uparrow$  & F1 $\uparrow$  \\
\midrule
ViT\cite{dosovitskiy2021image} & 80.82$\pm$0.86\% &  80.76$\pm$ 0.87 \% & 80.75$\pm$ 0.86 \%  \\ 
ViT\cite{dosovitskiy2021image} + S & 80.24$\pm$ 0.47\%  &  80.49$\pm$ 0.63\% &  80.44$\pm$ 0.57\% \\
ViT\cite{dosovitskiy2021image} + LC &  \underline{81.33$\pm$ 0.23\%}  & \underline{81.29$\pm$ 0.22\%}  & \underline{81.30$\pm$ 0.23\%}  \\
ViT\cite{dosovitskiy2021image} + LP + rand & 59.43 $\pm$ 1.21 \%  & 59.56 $\pm$1.32 \% & 59.65 $\pm$1.29\%  \\
ViT\cite{dosovitskiy2021image} + LP &  79.51$\pm$ 0.23\% & 79.37$\pm$ 0.34\% & 79.37$\pm$ 0.35\%    \\ 
ViT\cite{dosovitskiy2021image} + LP + Reg & \textbf{81.57$\pm$ 0.29\%} &\textbf{81.53$\pm$ 0.30\%}  &  \textbf{81.51$\pm$ 0.30\%} \\
\bottomrule
\end{tabular}
\vspace{1mm}
\caption{\textbf{Cifar-10-INRs Classification task.} Classification results on our CIFAR-INR dataset with different grouping methods. Abbreviations are introduced in Figure~\ref{fig:grouping}. All results are averaged over 5 runs. Baseline results align with results reported using \href{https://github.com/kentaroy47/vision-transformers-cifar10.git}{CIFAR-10 images}.
}
\label{tab:cifar_main_results}
\end{table}

%
%



\begin{table}
\centering
\begin{minipage}{\linewidth}
\centering
\begin{tabular}{l|ccll}
\toprule
Method & Params & GFLOPs & mIOU $\uparrow$(fine)  & mIOU $\uparrow$(coarse) \\
\midrule
MiT-B0\cite{xie2021segformer} MiT-B0 & 3.7  & 31.5 & 39.95 $\pm$0.9 & 42.67 $\pm$0.8   \\
MiT-B0\cite{xie2021segformer}+LC & 3.7 & 112.5 & \underline{40.33$\pm$0.7} & \underline{42.70 $\pm$0.6} \\
MiT-B0\cite{xie2021segformer}+LP+Reg & 4.0 & 112.5 & \textbf{40.61$\pm$0.6} & \textbf{43.29 $\pm$0.3}  \\
\bottomrule
\end{tabular}
\vspace{1mm}
\caption{\textbf{CityScapes-F\cite{cordts2016cityscapes} Segmentation.} Results obtained on the Cityscapes fine annotation dataset.
}
\label{tab:cityscapes_results}
\end{minipage}\hfill

\end{table}

\section{The Implicit-Zoo Benchmark and Experimental Results}\label{benchmark}

We report four benchmarks on the Implicit-Zoo Dataset, covering image classification, semantic segmentation, and 3D pose regression tasks. Our focus is on the RGB input space. Additional investigations on methods that use only the weight space and also training details can be found in the supplementary materials.

\subsection{Benchmark methods} We choose Vision Transformer and its variants as main method for its state of the art performance across multiple tasks and to demonstrate the effectiveness of the learnable tokenizer. For classification we choose \cite{dosovitskiy2021image} as baseline method and for segmentation we use Segformer\cite{xie2021segformer} for benchmark approach for its efficiency and light-weight design. For pose regression we use method shown in \ref{fig:pose_regressor}.

\subsection{Benchmark experiments and results}
\textbf{CIFAR-10-INRs Classification:} We train ViT classifier from scratch with batch size 512 for 200 epochs. We compare five different grouping methods with the baseline. In addition to the approaches mentioned in Sec. \ref{rgb_grouping}, we found that if all pixels are allowed to move without any constraints, multiple pixels may converge to the closing location due to local minima shown in Fig.\ref{fig:vis_group_vit_free}. To address this, we introduced a regularization term to penalize too close coordinates within the same token $t$ as shown in Eqn.\ref{eqn:regularization}. We abbreviate it as "LP+reg". For N coordinates $\{x_i\}_{i=1}^N$ within that token, we use $\mathcal{L}1$ gate loss and choose the threshold $\alpha$ to be $ min(1/H,1/W)$ where H, W is the height and width of input images, such that,
\begin{equation}\label{eqn:regularization}
    \mathcal{L}_{reg} = \sum_{i=1}^{N} \sum_{j=1, j \neq i}^{N} \text{ReLU}\left(\alpha - \|{x}_i - {x}_j\|_2\right).
\end{equation}
 Implementing this approach reveals that the proposed ``LC" and ``LP+reg" methods surprisingly outperform the baseline by 0.51 $\%$ and 0.75 $\%$ in accuracy. Conversely, the "LP+rand" method performs poorly, which is consistent with our expectations due to the loss of local geometry and appearance information \cite{qian2022makes}. As illustrated in Fig. \ref{fig:vis_moving_center_patch}, boundary patches tend to move towards the center, while center patches tend to overlap. With Non-Close regularization, the coordinates, as shown in Fig. \ref{fig:vis_reg}, retain a grid structure. This helps preserve local geometry information while leave patches the flexibility to grow and overlap.

\textbf{ImageNet-100 INRs Classification:}\label{imagenet_exps} We conduct fine-tuning experiment similar to \cite{dosovitskiy2021image} and use the best RGB grouping approaches from previous experiments. After fine-tuning 8000 steps based on re-finetuned from augreg 21k-1k\cite{qian2022makes}, we report similar performance improve in accuracy in Tab. \ref{tab:imagenet100_results}, It is minimal potentially because the pre-trained ViT is based on fixed patches. Additionally, "LC" performs better "LP", which could also be attributed to the pre-training on grid-like tokenization.

\begin{table}[h]
\centering
\begin{minipage}{\linewidth}
\centering
\begin{tabular}{l|ccc}
\toprule
Method & Acc $\uparrow$ & Pre $\uparrow$  & F1 $\uparrow$  \\
\midrule
ViT\cite{dosovitskiy2021image} & 84.81$\pm$0.92 \% & 85.2$\pm$0.87  \%& 84.58$\pm$1.05  \%\\
ViT\cite{dosovitskiy2021image}+LC & \textbf{85.02$\pm$1.02} \% & \underline{85.28$\pm$1.01}  \% & \underline{84.64$\pm$1.03  \%}  \\
ViT\cite{dosovitskiy2021image}+LP+Reg & \underline{84.92$\pm$0.93}   \%& \textbf{85.30$\pm$0.91}  \% & \textbf{84.74$\pm$1.05\%} \\
\bottomrule
\end{tabular}
\vspace{1mm}
\caption{\textbf{ImageNet-100\cite{tian2020contrastive} Classification task.} Fine-tuning experiments on the ImageNet-100 dataset yield results similar to those observed in the CIFAR-10 experiment \ref{tab:cifar_main_results}. The improved performance is attributed to the proposed learnable tokenization, thanks to the Implicit-Zoo dataset.
}\label{tab:imagenet100_results}
\end{minipage}%
\hfill

\end{table}


\textbf{CityScapes-INRs Semantic Segmentation:} Follow experiments in \cite{xie2021segformer} which use pretrained encoder MiT-B0 on ImageNet-1K and then fine-tune it on Cityscapes-INRs for 40k steps with batch size 16. We report also the Params and GFLOPs for using our learnable tokenizers. Note that optimizing tokenizer ffor all coordinates in high-resolution images can be very computationally expensive in terms of GFLOPs. We report baselines results similar to \cite{xie2021segformer}. Surprisingly, using learnable tokenizer also improves performance in pixel-to-pixel tasks like segmentation. The misalignment between input pixels and the supervised pixels with labels can be can be resolved by tokenization and self-attention mechanisms, which help to locally aggregate relevant information.


\begin{table}[h]
\centering
\resizebox{0.8\linewidth}{!}{
\begin{tabular}{l|ll|ll|ll } 

\toprule
Metric &  Ours &  Ours+Pre & Ours+LC  & Ours+Pre+LC  & Ours+LP & Ours+Pre+LP \\
\midrule
\multicolumn{7}{c}{Seen Scenes}  \\
\midrule
TE(cm) $\downarrow$& 3.50  & 3.13 &  3.30 &  \underline{3.12} & 3.22 &  \textbf{2.99}\\
RE $\downarrow$& 15.29 \textdegree  & \underline{14.40} \textdegree &  15.67  \textdegree & 14.59 \textdegree  &  14.67\textdegree & \textbf{14.17\textdegree}  \\
RE@5 $\uparrow$& 44.36\% & 47.30\% & 43.55\%  & 47.34 \% & 45.11\% & \textbf{50.02 \%}   \\
RE@15 $\uparrow$& 75.30 \%& \underline{76.50\%}  &  74.95 \% & 76.41\% &  75.21 \% &  \textbf{76.60 \% } \\
RE@30 $\uparrow$& 83.35\%  & \underline{83.85 \%}  &   83.27 \% &  83.75 \% &   83.44 \% & \textbf{83.95 \%}  \\
RE+Ref $\downarrow$& 6.93 \textdegree  & 4.35 \textdegree  &  7.13 \textdegree  &   \underline{4.22 \textdegree}  & 4.07  \textdegree  & \textbf{3.91 \textdegree}    \\

\midrule
\multicolumn{7}{c}{Unseen Scenes}  \\
\midrule

TE(cm) $\downarrow$& 6.91  & 6.61  & 6.92  & \textbf{5.96} &  6.86 & \underline{5.99} \\
RE $\downarrow$& 21.90 \textdegree & 20.24 \textdegree & 21.83 \textdegree  &\underline{20.09}   \textdegree &  21.68 \textdegree &  \textbf{20.02 \textdegree}  \\
RE@5 $\uparrow$&  27.46\% & 29.94 \%  & 27.46\%  & \textbf{30.77\%}  &  27.97\% & \underline{30.19\%}  \\
RE@15 $\uparrow$& 64.19 \% & 66.65\%  & 64.20\% & \textbf{67.07\%}  &  64.23 \% & \underline{66.9 \%}  \\
RE@30 $\uparrow$& 77.50 \%& 79.42 \%  & 77.40\%  & \underline{79.44\%}   & 79.61\% &\textbf{79.75\%}  \\
RE+Ref $\downarrow$& 9.86 \textdegree  & \textbf{9.07} \textdegree  & 9.85 \textdegree   & 9.70 \textdegree   & 9.01 \textdegree  & \textbf{8.09 \textdegree} \\

\bottomrule
\end{tabular}}
\vspace{1mm}
\caption{\textbf{OmniObject3D-INRs pose regression.} We present pose regression results for both seen and unseen scenes in our Omniobject3D dataset. `Pre' denotes a pretrained encoder, `Ref' refers to further refinement. Our proposed learnable tokenization achieves better pose regression results.}
\label{tab:nerf_pose_estimation}
\end{table}

\begin{figure}
    \centering
    \begin{subfigure}[t]{0.25\textwidth}
        \includegraphics[angle=90, width=\textwidth, trim={0 8cm 20cm 10cm},clip]{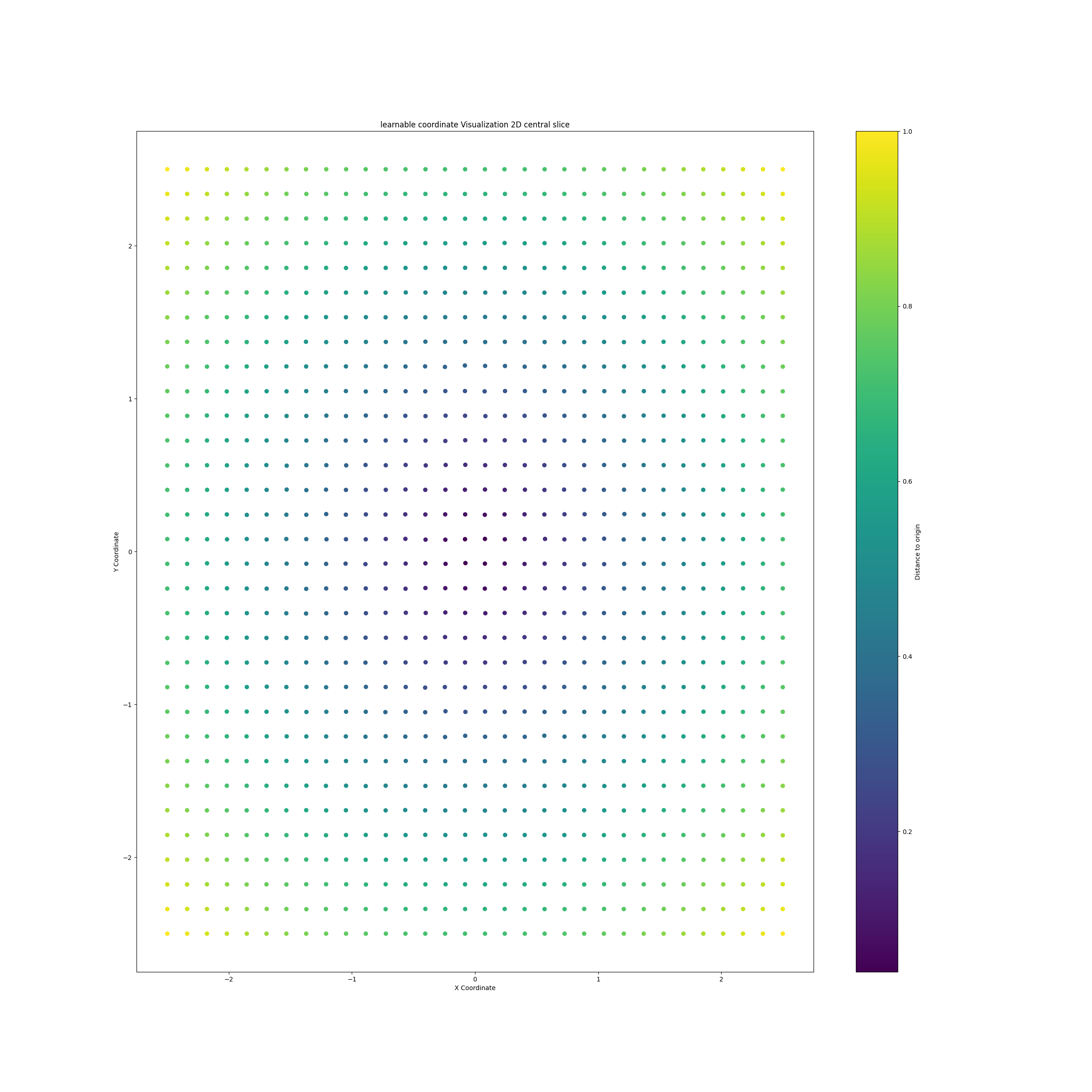}  
        \caption{Uniform Volumize}
        \label{fig:group_vit}
    \end{subfigure}
    \hfill
    \begin{subfigure}[t]{0.25\textwidth}
        \includegraphics[angle=90, width=\textwidth, trim={0 8cm 20cm 10cm},clip]{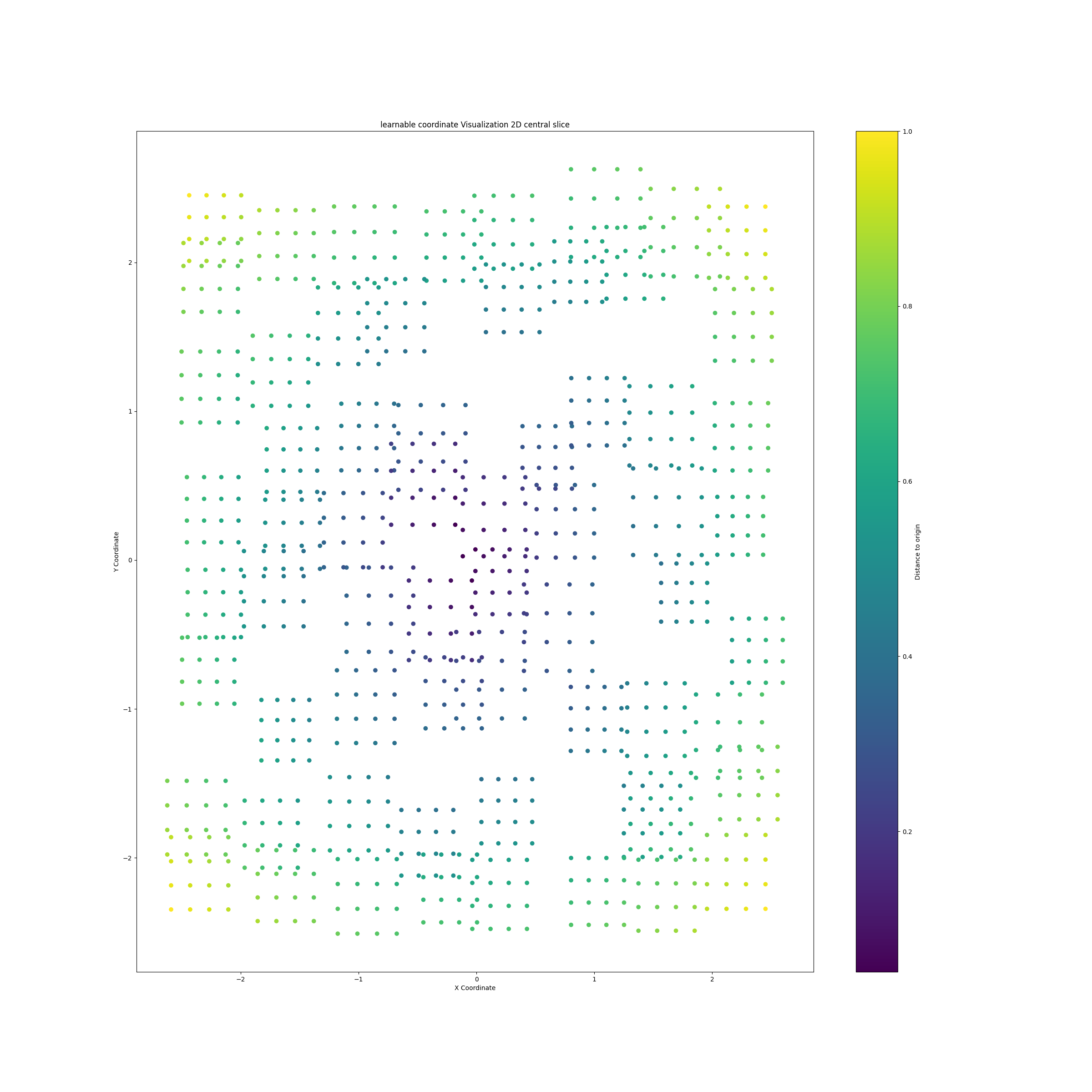}
        \caption{Learnable Volume Center + Learnable Scale}
        \label{fig:3d_lc}
    \end{subfigure}
    \hfill
    \begin{subfigure}[t]{0.25\textwidth}
        \includegraphics[angle=90, width=\textwidth, trim={0 8cm 20cm 10cm},clip]{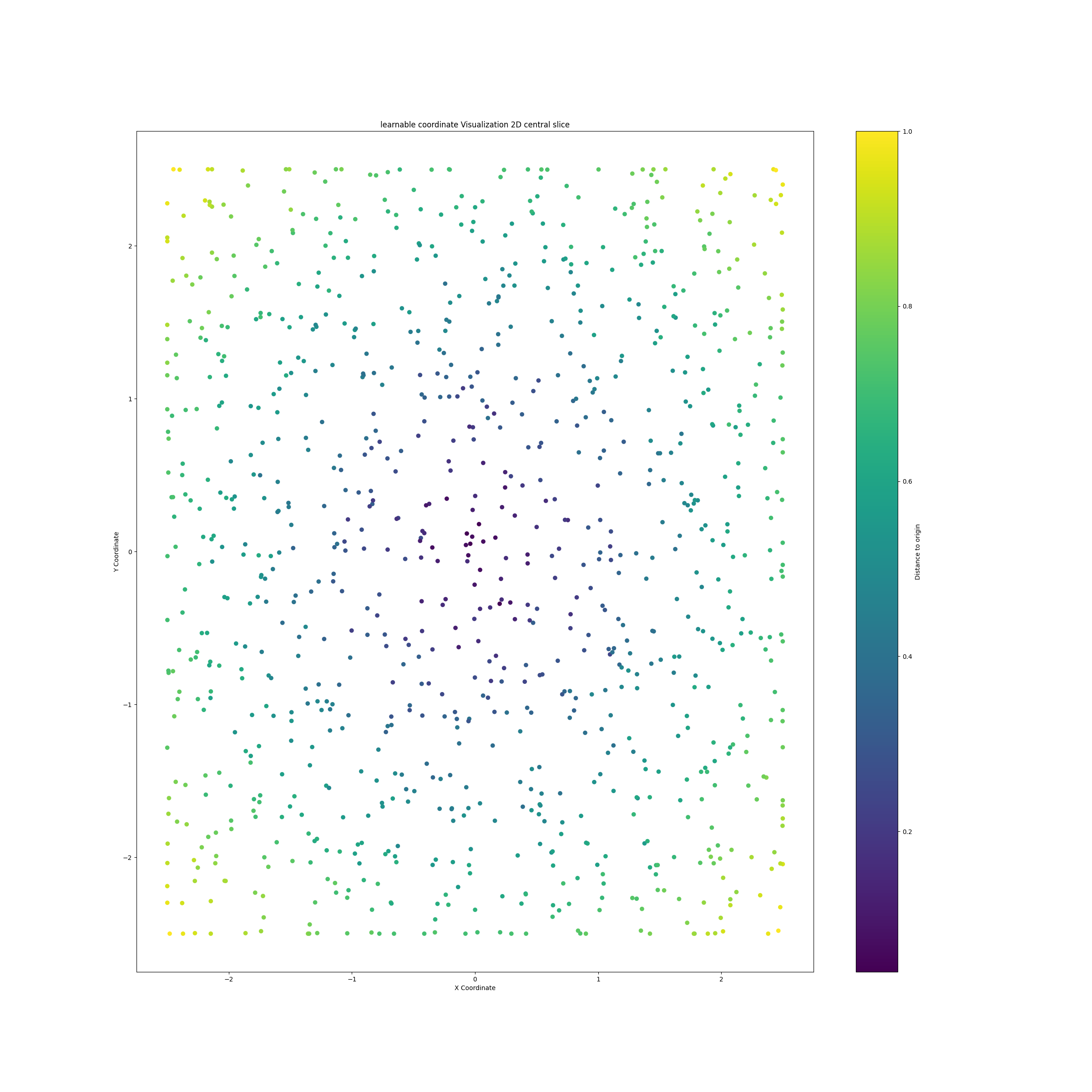}
        \caption{Learnable Points}
        \label{fig:3d_lp}
    \end{subfigure}
    \hfill
    \vspace{1em}
    \begin{subfigure}[t]{0.25\textwidth}
        \includegraphics[width=\textwidth, trim={2cm 0cm  2cm  5cm},clip]{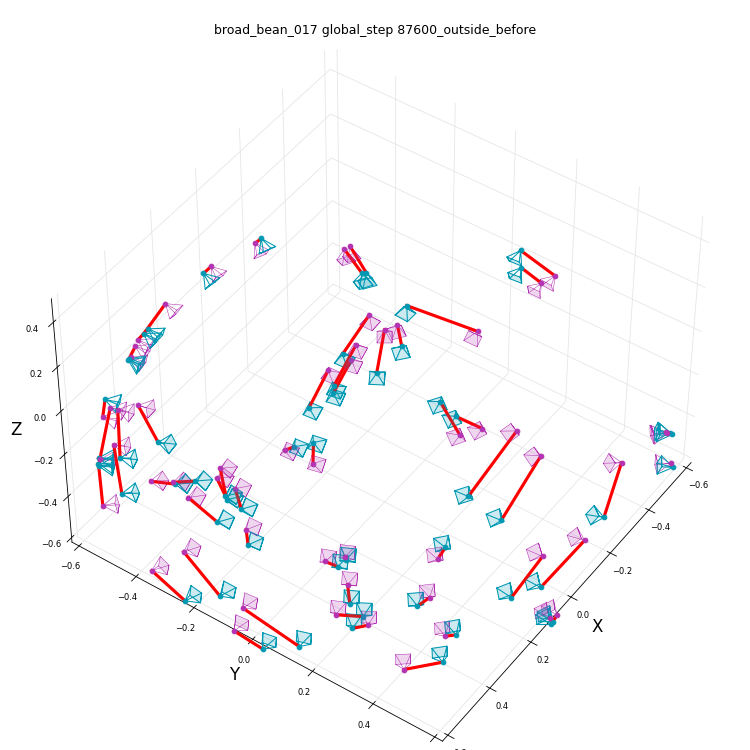}
        \caption{Coarse Pose }
        \label{fig:3d_coarse_vis}
    \end{subfigure}
    \hfill
    \begin{subfigure}[t]{0.25\textwidth}
        \includegraphics[width=\textwidth, trim={2cm 0cm  2cm  5cm},clip]{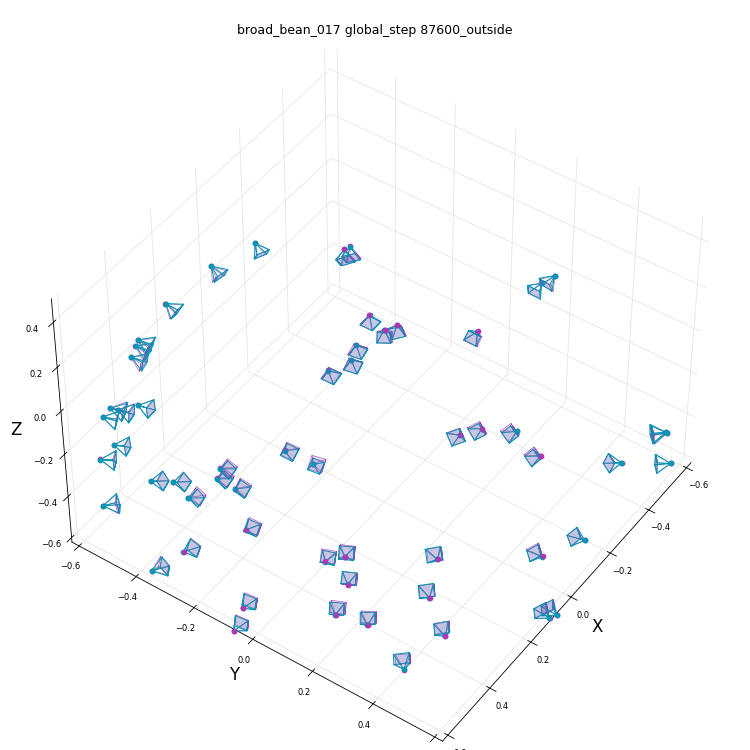}
        \caption{Refined Pose}
        \label{fig:3d_fine}
    \end{subfigure}
    \hfill
    \begin{subfigure}[t]{0.25\textwidth}
        \includegraphics[width=\textwidth, trim={2cm 0cm  2cm  5cm},clip]{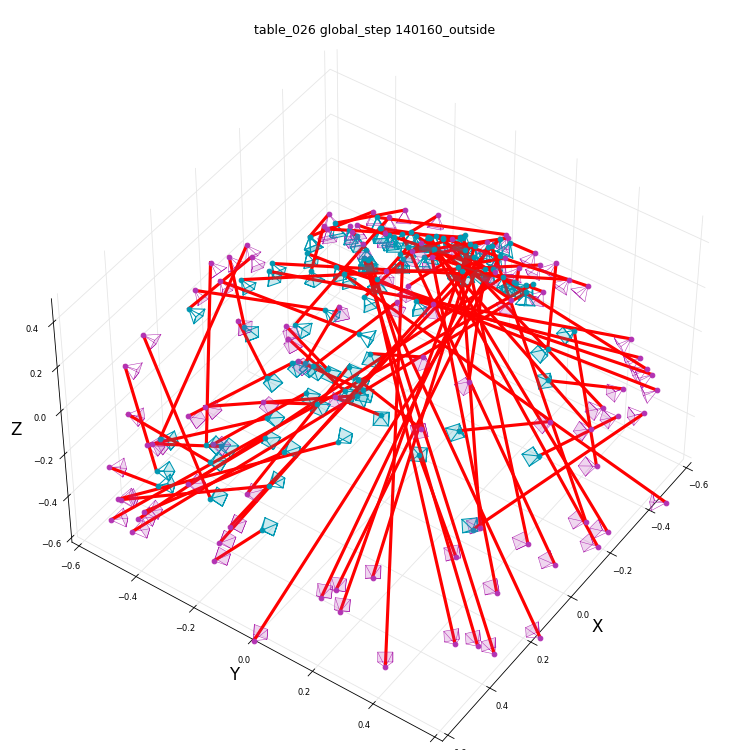}
        \caption{Failure case}
        \label{fig:failiure_case}
    \end{subfigure}
    \hfill


    \caption{\textbf{3D learnable token and pose regression.} We visualize the center slices of the learnable 3D tokenizer (top), where the colors indicate initialized positions. The coarse poses are directly regressed using the network trained on the Implicit-Zoo dataset. Red lines show the translation errors.}
    \label{fig:3d_grid}
    
\end{figure}

\textbf{Omniobject3D-INRs pose regression:} Our proposed small model discretize scenes to a 32x32x32 volume and volume points are queried to and get rgb and density. We then concatenate them together and tokenize it to volume token with 3D convolutions shown in \ref{fig:pose_regressor}.We compute the $\mathcal{L}2$ distance for translation and for rotational error we use shortest rotation distance. Moreover we also report RE$@\beta$ where $\beta \in ({5,15,30})$ which indicated the percentage of rotational error below a threshold $\beta$. Our findings show that pre-trained volume encoder improves results across all methods with different grouping strategies. Moreover, optimizing the 3D tokenizer yields better performance. Although the improvements are modest, qualitative images \ref{fig:3d_lc} reveal that the learned patches exhibit large overlap and leave blank areas. This suggests the potential to reduce the number of patches. Additionally, our method identifies coarse positions close to the ground truth pose, which can be further optimized using other pose regression methods \cite{lin2021barf,yen2021inerf}. More details regarding model and pre-training can be find in supplementary.

\section{Conclusion and Limitations}\label{limitation}
In this paper, we introduced \textit{Implicit-Zoo}, a large-scale implicit function dataset developed over nearly 1000 GPU days. Through strict data quality check, we have ensured its high quality. Using this dataset, we benchmarked a range of tasks, including 2D classification and segmentation. Additionally, we proposed a transformer-based approach for pose regression utilizing trained 3D neural radiance fields. By incorporating a learnable tokenizer, we enhanced the benchmark methods and discovered valuable insights for future in tokenizer research. Our work has few limitations: Scalability is restricted due to INR querying, resulting in small batch and model sizes in benchmark experiments, which limits from-scratch training of complex models. We set a PSNR threshold of 30 for dataset expansion, but this may cause artifacts in repetitive backgrounds, thus requires additional data refinement. Lastly, our pose regression task struggles with symmetric objects (Fig. \ref{fig:failiure_case}); this may be addressed using symmetry-aware representations \cite{hodan2020epos}. We believe our dataset and its showcased applications open up new avenues for the future research.

\title{Implicit Zoo: A Large-Scale Dataset of Neural Implicit Functions for 2D and 3D Scene \\ Supplementary Material}





\section{Supplementary Material}

In this supplementary material, we first detail the data generation process in Section~\ref{sec_data} and provide more information on the implementation of the learnable tokenizer in Section~\ref{sec_token}. Next, we present additional details and experimental results on the benchmarks in Section~\ref{sec_benchmark}. Finally, we provide information needed in checklist in Section~\ref{sec_checklist}. To gain a better understanding of our dataset and proposed benchmarks, please refer to the introductory video in the supplementary materials or the one available on our \hyperlink{https://github.com/qimaqi/Implicit-Zoo}{project page}, which offers an overview of our dataset and its applications.

\subsection{Additional details of Dataset Generation}\label{sec_data}

\begin{figure}[h]
\centering
        \centering
        \includegraphics[width=\linewidth, trim={0  0  0  0},clip ]{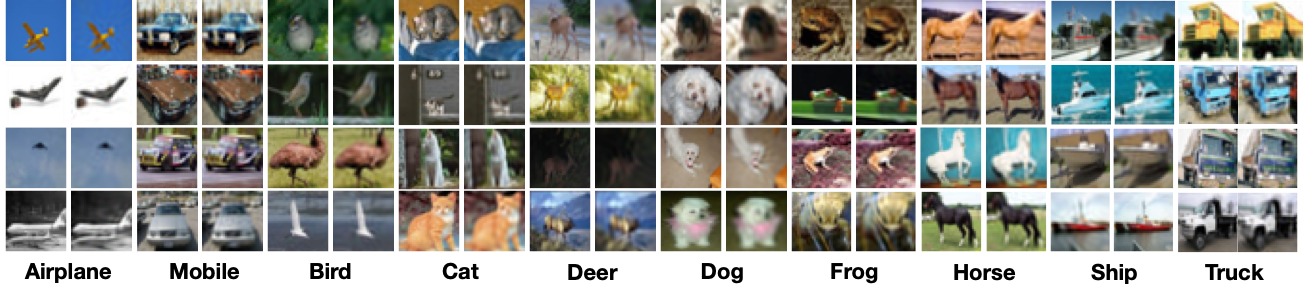}
\caption{\textbf{Additional examples on CIFAR-10-INRs dataset} We present additional data examples, where the left side of each image pair shows the ground truth and the right side displays the results queried from the INRs.
\label{fig:cifar_add_example}}
\end{figure}

\begin{figure}[h]
\centering
        \centering
        \includegraphics[width=\linewidth, trim={0  0  0  0},clip ]{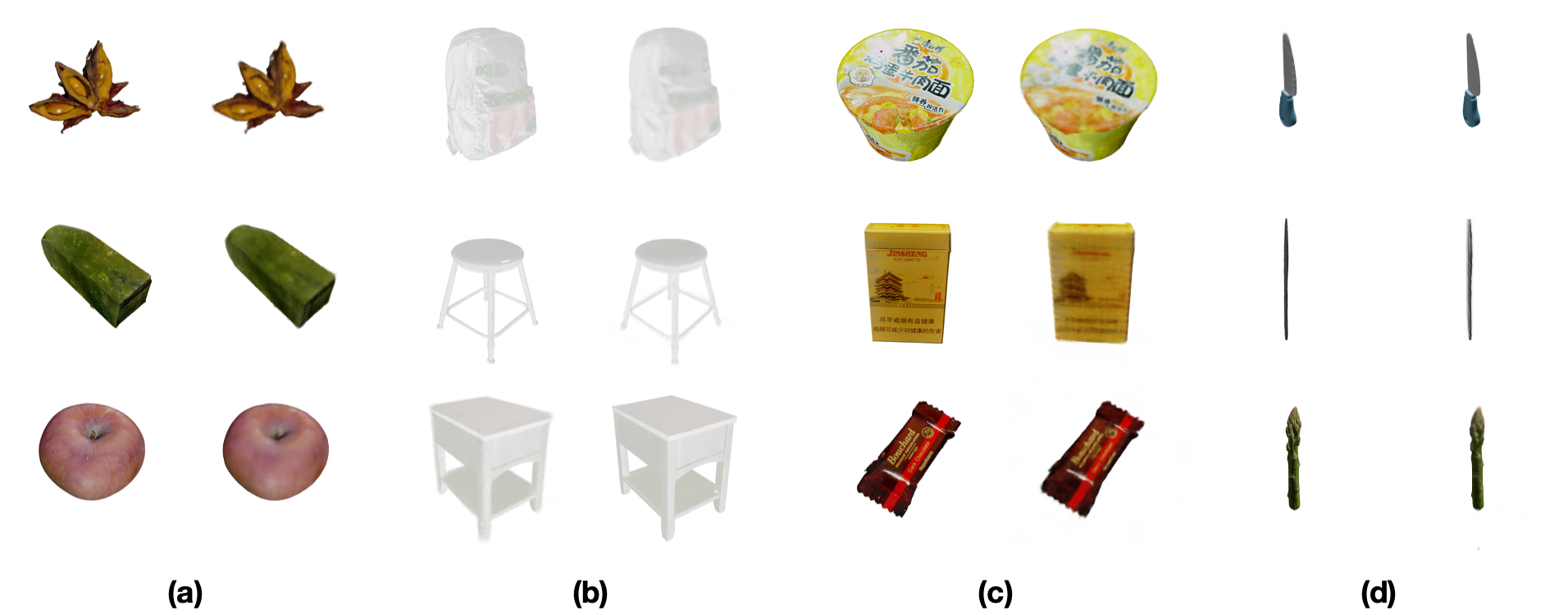}
\caption{\textbf{Additional examples on Omniobject3D dataset} We present additional examples on Omniobjecct3D. We observe that when objects are large, have rich colors, and relatively simple surfaces, our reconstruction performs very well (a). However, in more challenging cases such as (b) shallow-colored objects, (c) complex surfaces with text information, and (d) small or thin objects, the reconstruction quality is less satisfactory.
\label{fig:omniobject3d_add_example}}
\end{figure}

\textbf{Speed up Training} 
In \cite{dupont2022data}, the authors propose meta-learning and implicit function modulation to accelerate the training process. Similarly, \cite{erkocc2023hyperdiffusion} and \cite{deluigi2023deep} reduce training time by employing smaller models and optimizing the number of iterations. In our 2D dataset, we observed that normalizing images before training implicit functions significantly speeds up training iterations. Therefore, we chose to normalize images and get rid of Sigmoid activation function in the final layer. In the 3D cases, we use a small model with 4 layers and a width of 128, without any skip connections. During training to enhance the performance with limited iterations, we propose an adaptive sampling method that focuses more on rays corresponding to 2D RGB values that are not white (likely to be the background). This approach is particularly beneficial for handling light-colored cases and tiny objects as shown in Fig ~\ref{fig:omniobject3d_add_example}.
We observed that the training loss converged at 20k steps with learning rate 5e-4.

\begin{figure}[ht]
\centering
        \centering
        \includegraphics[width=\linewidth, trim={0  0  0  0},clip ]{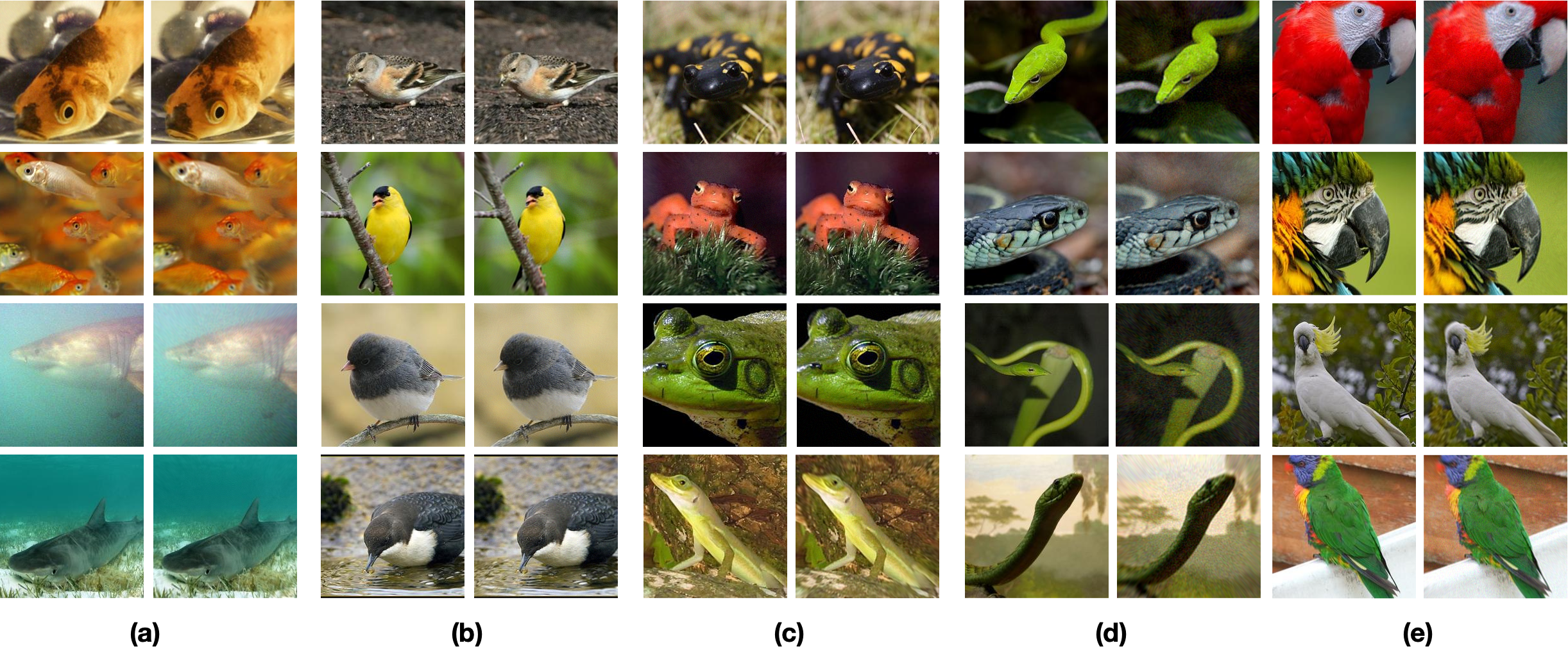}
\caption{\textbf{Additional examples on ImageNet dataset} We present additional animal images from the ImageNet dataset, which is one of the motivations behind naming this work \textit{Implicit Zoo}. Comparing with the ground truth images on the left, the reconstructions are of very high quality.
\label{fig:image_add_example}}
\end{figure}

\begin{figure}
\centering
        \centering
        \includegraphics[width=\linewidth, trim={0  0  0  0},clip ]{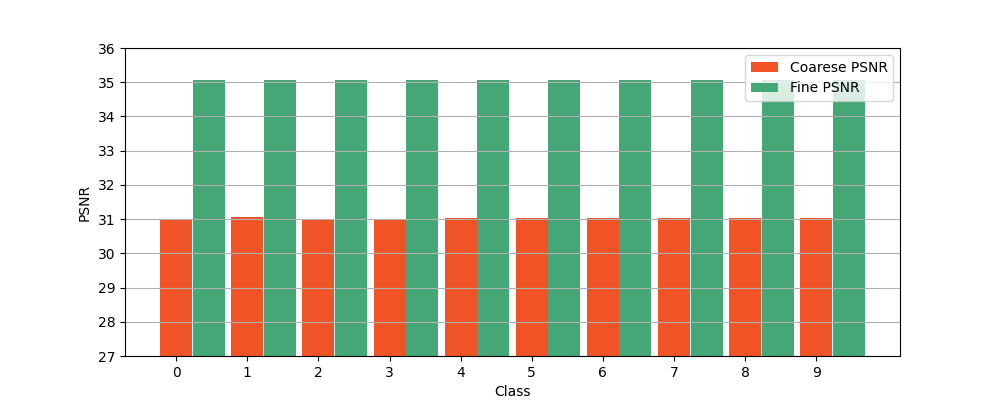}
\caption{\textbf{PSNR for each class in CIFAR-10-INRs} We report the PSNR for all classes of CIFAR-10 INRs. Before refinement, the standard deviation across different tasks is 0.013, and after refinement, it is 0.005.
\label{fig:cifar_psnr}}
\end{figure}

\textbf{More Examples of data}
We provide more examples of data on ~\ref{fig:cifar_add_example}, ~\ref{fig:image_add_example}, ~\ref{fig:cifar_add_example}, ~\ref{fig:omniobject3d_add_example}. Note that if you zoom in, you may notice some artifacts in the CIFAR-10 dataset. For example, in the bottom row of the dog category in Fig~\ref{fig:cifar_add_example}
, the dogs appear slightly blurry. For a 32x32 image, a PSNR of 30 results in more noticeable visual differences compared to larger images, as seen in Fig~\ref{fig:image_add_example}. To address this issue, we refined the CIFAR-10 data as described in the main paper, increasing the average PSNR to approximately 35 and resulting in a smaller standard deviation across different classes, as shown in Fig~\ref{fig:cifar_psnr}. Additional experiments on the refined dataset, reported in Table~\ref{tab:cifar_add_results}, align with the findings in the main paper.

\begin{figure}[h]
\centering
        \centering
        \includegraphics[width=\linewidth, trim={0  0  0  0},clip ]{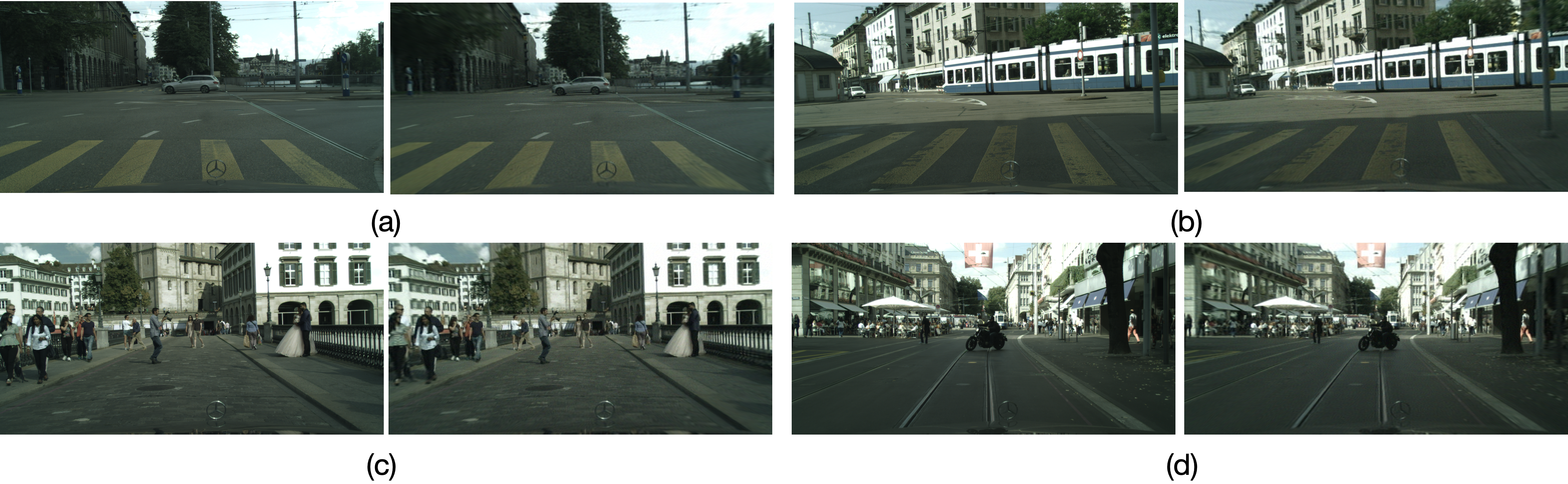}
\caption{\textbf{Additional examples on Cityscapes dataset} We present additional data samples from Cityscapes-INRs. Notably, fine details such as pedestrians in (c) and significant illumination changes in (d) are well-preserved in the reconstructions.
\label{fig:cityscapes_add_example}}
\end{figure}

\begin{figure}
\centering
        \centering
        \includegraphics[width=\linewidth, trim={0  0  0  0},clip ]{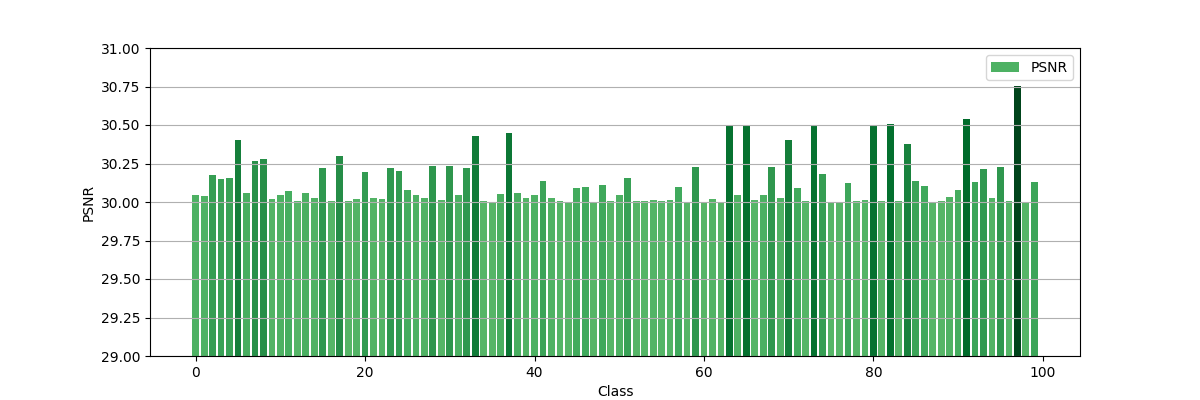}
\caption{\textbf{PSNR for each class in ImageNet-100-INRs} We report PSNR on ImageNet-100 with standard deviation 0.157.
\label{fig:image_psnr}}
\end{figure}

\textbf{Data statistics} We report PSNR across classes of dataset in Fig ~\ref{fig:cifar_psnr}, ~\ref{fig:image_psnr}. For Cityscapes-INRs results please refer to main page. Note that the PSNR differences in 2D cases are minimal due to the quality control and further refinement we implemented. Some of the classes in ImageNet results are higher than others, indicating better performance achieved during the initial phase of training.

\begin{figure}
\centering
        \centering
        \includegraphics[width=\linewidth, trim={0  0  0  0},clip ]{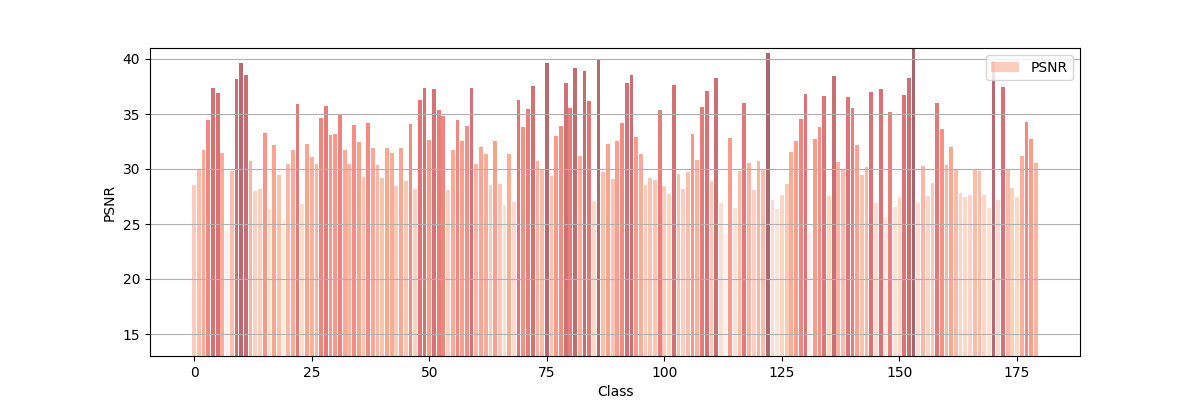}
\caption{\textbf{PSNR for each class in Omniobject3D-INRs-INRs} We report PSNR on Omniobject3D with standard deviation 3.87. 
\label{fig:omni_psnr}}
\end{figure}

\textbf{Scene filtering} As shown in Fig ~\ref{fig:omni_psnr} the rendering PSNR for novel view changes a lot cross different classes (180 claases) with standard deviation 3.87. This is mainly because the various objects the dataset include. As shown in Fig ~\ref{fig:omniobject3d_add_example},
we observe that PSNR tends to be higher for light-colored objects because their colors align with the white-background assumption \cite{mildenhall2021nerf}. A similar trend is observed for small objects. During data filtering, we first exclude cases with a PSNR below 25. For classes with fewer than five scenes, we ignore the entire scene in this class. Ultimately, we retain 5,287 valid scenes for our experiments.

\textbf{Data releasing}. We uploaded CIFAR-10-INRs, ImageNet-10-INRs Omniobject3D to Kaggle and can be found in \hyperlink{https://github.com/qimaqi/Implicit-Zoo}{project page}. The Cityscapes-INRs dataset will be released shortly on the Cityscapes team's official \href{https://www.cityscapes-dataset.com/} after this paper is published publicly. Additionally, we are working on a refined version of ImageNet with PSNR > 35 and training a larger NeRF model on Omniobject3D, utilizing a coarse and fine model with 8 layers and a width of 256. The benchmark code will also be released on the above webpage.


\begin{figure}[h]
\centering
        \centering
        \includegraphics[width=0.9\linewidth, trim={0  0  0  0},clip ]{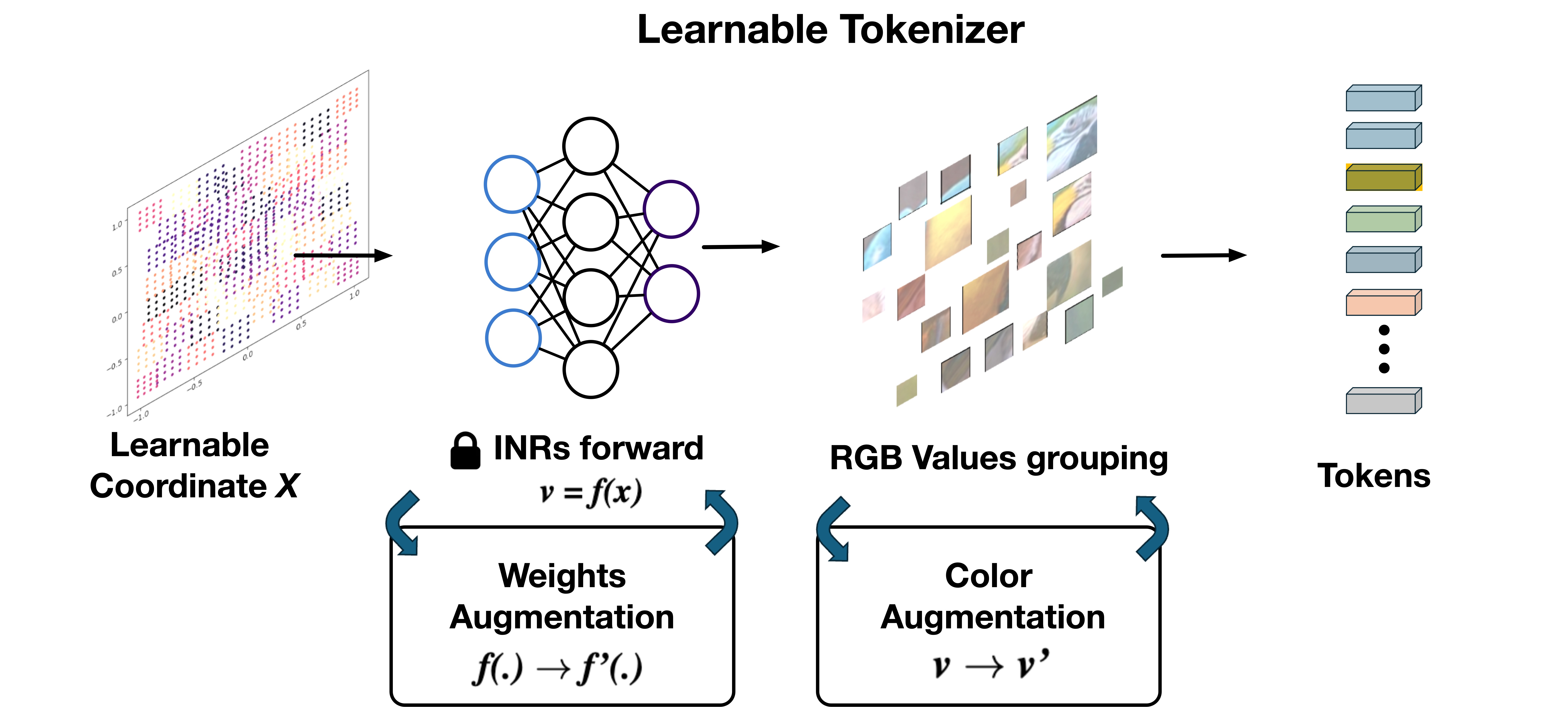}
\caption{\textbf{Differential Augmentation} We propose geometric augmentation in weights-space and color augmentation in RGB-space.Following \cite{cubuk2019randaugment} we propose 15 differentiable transforms which enhance our dataset application.
\label{fig:augmentation_pipeline}}
\end{figure}

\subsection{Additional details of Learnable tokenizer}\label{sec_token}

\textbf{2D implementation} We provide more detailed information on learnable tokenizer and different RGB grouping on 2D implementation here. We first map the coordinates to $(-1,1)$ and then divide them uniformly with patch size $P$ to $N$ patches, each containing $P^2$ coordinates. We calculate the center coordinate $c_i$ for each patch $i \in \{1,2,...N\}$ and determine the coordinate difference of each coordinates to the center coordinate $d_{ij}$, where $j \in \{1,2,...P^2\}$. Thus, all coordinates $x_{ij}$ can expressed as $x_{ij}=c_i + d_{ij}$. For (b) Learnable Scaling, we introduce a learnable scaling factor $s_i$ for each patch. The queried coordinate is then given by $x_{ij}' = c_i + s_id_{ij}$. For (c) Learnable Centers we make $c_i$ themselves also learnable. Both method (b) and (c) keep the grid shape. For (d) Learnable pixels instead of learning a coordinate difference we directly make all coordinate $x_ij$ learnable. Finally in (d) we divide at beginning the coordinates randomly.Furthermore, to stabilize the training and ensure the learnable scale remains non-negative and the learnable pixels stay within the range $(-1,1)$ a extra $Tanh(.)$ activation is applied on scaling factor and $Sin(.)$ is added on learnable coordaintes.

\textbf{Differential augmentation} As discussed in the main paper, differential augmentation is crucial to ensure that gradients can backpropagate to the learnable tokenizer. Unlike previous works \cite{zhou2024neural, zhou2023permutation} hat train more INRs on augmented data, we implement \cite{cubuk2019randaugment} in a differential manner. Note that we propose to implement non-differential geometric augmentations in weight-space to make them differentiable. As shown in Fig ~\ref{fig:augmentation_pipeline}, we first implement geometric augmentation such as Rotate, ShearX, ShearY, TranslateX, TranslateY, Cutout in by modifying the first layer of INRs. Specifically, we adjust the weights and biases as $W' = W + W_t$ and $b' = b + b_t$. This is calculated by $W'(x')+b' = Wx + b$ where $x'$ is the corresponding coordinates after transformation. 

\begin{figure}[h]
\centering
        \centering
        \includegraphics[width=0.9\linewidth, trim={0  0  0  0},clip ]{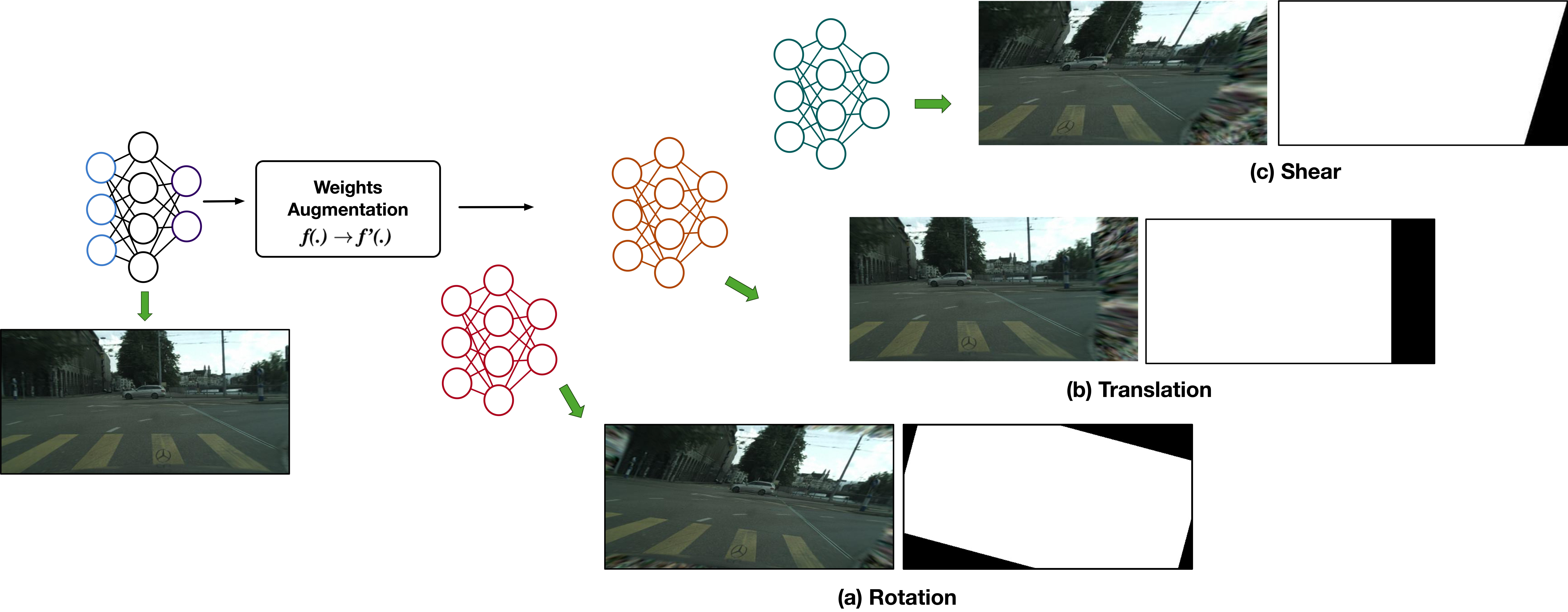}
\caption{\textbf{Masking out-of-range values}: We implement zero-padding by applying same geometric transformation for a binary mask and we show examples for (a)rotation, (b)translation and (c)shear operation. Out-of-range RGB values are clearly visible in the bottom right corner of the transformed image.
\label{fig:masked_augmentation}}
\end{figure}

Next, we implement color augmentations such as AutoContrast, Equalize, Solarize, Color Balance, Invert, Contrast, Brightness, and Sharpness in RGB space. Two main challenges arise: first, some color augmentations are non-differentiable, such as the Equalize operation, which creates a uniform distribution of grayscale values in the output image, and the Posterize operation, which reduces the number of bits for each color channel. To address that we first calculate the transform $T$ outcome of these two operation and add the residual to our rgb value. $\triangle v = T(v) - v$ and $v' = v + \triangle v$. Note that we do not apply this residual addition to all color transformations because we want proposed learnable tokenizer to remain learn from color augmentations. Secondly some out-of-range RGB values can appear due to geometric augmentations $x' \notin (-1,1)$, as illustrated in the Fig~\ref{fig:masked_augmentation}. These values can significantly impact downstream tasks such as segmentation and can also interfere with RGB augmentations process. To address this, we propose applying the same geometric transformations to a binary mask $M$ with zero padding for out-of-range values. Before performing Color augmentations, we first apply the mask operation $v_{mask}=M v$. We adopt some operations from \cite{eriba2019kornia}.

\begin{table}[t]
\centering
\begin{tabular}{l|ccc}
Augmentation &  Rotate & Translate & ShearX \\
\toprule

Implementation &  $W_t$ = $\begin{bmatrix} cos(\theta) & -sin(\theta) \\ sin(\theta) &  cos(\theta) \end{bmatrix}$   &  $ b_t =  W \triangle b $  &    $W_t$ = $\begin{bmatrix} 1 & s \\ 0 &  1 \end{bmatrix}$,   $ b_t =  W$ $\begin{bmatrix} s \\ 0  \end{bmatrix}$   \\

\midrule
\end{tabular}
\caption{\textbf{Implantation of geometric augmentation on weight-space}: We implement geometric transformation by modifying the weight $W$ and bias $b$ in first layer of INRs}
\label{tab:cifar_main_results}
\end{table}

\textbf{3D implementation} To lift up the learnable tokenizer to 3D volume tokenization we make following modification. Firstly we uniformly divide the space to $N$ volumes and each include $P^3$ 3D sample points. Then similarly for Learnable Centers + Learnable Scale we calculate the center points $c_i$ for each volume $i \in \{1,2,...N\}$. 
The remaining operations are similar to 2D process.


\begin{figure}[h]
\centering
        \centering
        \includegraphics[width=\linewidth, trim={0  0  0  0},clip ]{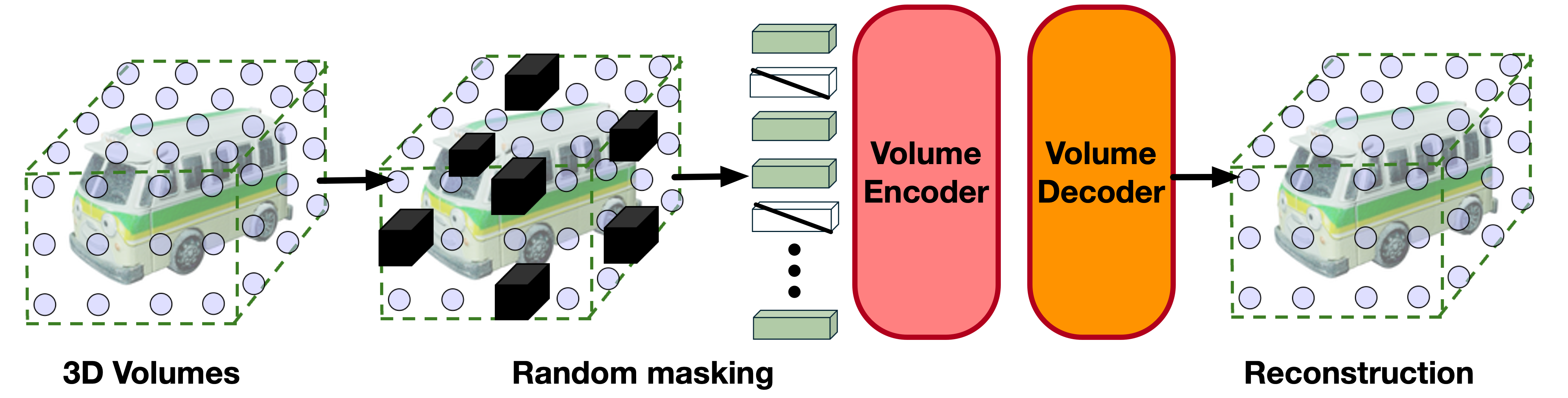}
\caption{\textbf{Illustration of Proposed 3D Volume Encoder pretraining mechanism} We randomly mask out 80\% of the volume tokens, allowing the encoder to operate only on the visible tokens. A small decoder processes the full set of encoded visible patches and masked tokens to reconstruct the input color volume and density volume. For qualitative results, please refer to Fig ~\ref{fig:results_mae}
\label{fig:pretrain_pipeline}}
\end{figure}

\subsection{Additional details of Benchmark}\label{sec_benchmark}

We provide detailed information on the implementation of our experiments and benchmarks. Additionally, we introduce our proposed pretraining mechanism, which has proven to be effective, as demonstrated in the main paper.

\subsubsection{Implementation details}
\textbf{CIFAR-10-INRs} We use ViT-tiny \cite{wolf2020huggingfaces} with a depth of 12 and 3 heads for multi-head self-attention. The patch size is set to 4, and we employ a convolution layer with a kernel size of 4 as the embedding layer. For optimization we use Adam optimizer \cite{kingma2017adam} with learning rate is 0.0001 for classify model and learnable tokenizer. Additionally, we apply CosineAnnealing leraning rate scheduler \cite{loshchilov2017sgdr}. For regularization,we use regularize weight of $w_{reg}=1$. 

 \textbf{ImageNet-100-INRs}: We utilize ViT-base \cite{dosovitskiy2021image,wolf2020huggingfaces} with pretrained model 21k-1k and fine-tuned on our ImageNet-100 dataset.Our experiments are conducted with a batch size of 32, using distributed training on GeForce RTX 4090 GPUs. We employ the AdamW optimizer ~\cite{loshchilov2019decoupled} with a learning rate of 1e-5 for the main model and the learnable tokenizer. A WarmupCosineAnnealingLR scheduler with one warmup epoch is used for learning rate adjustment. 

 \textbf{Cityscapes-INRs}: Similar to above we use AdamW optimizer ~\cite{loshchilov2019decoupled} with learning rate 1e-4 for segmentation model and 1e-5 for learnable token. This is because in segmentation task we do not want misalignment between supervision area and tokenizaiton area too large. Following \cite{xie2021segformer} we use PolynomialLR scheduler with power of 1.0.

 \textbf{Omniobject3D-INRs}: We first introduce the pretraining mechanism, by following \cite{feichtenhofer2022masked} we random masking the volume tokens as shown in Fig \ref{fig:pretrain_pipeline}. Unlike \cite{irshad2024nerfmae} we follow the standard ViT structure and focus on pose regression task. The volume encoder operates only on the unmasked tokens. The proposed volume encoder operates only on the unmasked tokens and consists of 12 layers, each with 3 heads and an embedding feature dimension of 192. For the decoding process, we utilize 8 layers transformer-based decoder with same head numbers and embedding dimension as encoder. We use a shared and learnable masked token to fill in the originally masked-out positions and apply the positional encoding of the original tokens. The decoder then predicts the original RGB and density values, using mean squared error (MSE) as the loss function.
 
Next, we use the pretrained encoder for INRs pose regression, demonstrating its effectiveness with improved results across all proposed learnable tokenizers. We train the model for 100 epochs with a batch size of 8, where each batch includes 24 sampled views of a given scene. For non-pretrained experiments, we use a learning rate of 1e-4, while for the pretrained volume encoder, we use a learning rate of 1e-5.

\subsubsection{Additional experiments}
We conducted weight-space-only experiments and additional experiments on refined CIFAR-10-INRs, training for 500 epochs. We selected DWSNet \cite{navon2023equivariant} and HyperRepresentation \cite{schürholt2022hyperrepresentations} for benchmarking, as they represent two primary approaches: one proposes a permutation-invariant network structure to process trained INRs, while the other learn strong encoder to tokenizes the network weights to latent feature.

For DWSNet, we used a 4-layer model with a hidden dimension of 64. For HyperRepresentation, we employed an 8-layer transformer with a hidden dimension of 512. Notably, DWSNet performed poorly on the CIFAR-10 experiments, likely due to the lack of additional information from the input coordinate domain for the RGB 3D higher dimension output \cite{navon2023equivariant}. HyperRepresentation performed better but still yielded unsatisfactory results compared to other RGB space-based methods.

\begin{figure}[h]
\centering
        \centering
        \includegraphics[width=0.9\linewidth, trim={0  0  0  0},clip ]{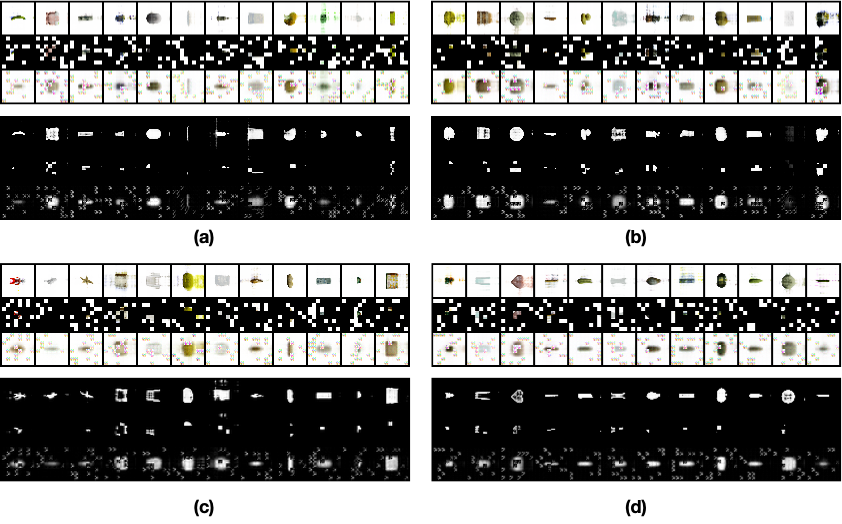}
\caption{\textbf{Visualization on validation set of Omniobject3D reconstruction} We present four batches of reconstruction results in panels (a), (b), (c), and (d). The top row displays the input RGB volumes and density volumes. The middle row shows the masked volumes, while the bottom row illustrates our reconstruction results. Each sample contains 32x32x32 sampled points, and with a volume size of 4, it generates 512 tokens, of which only 102 are visible. Note that the output in known patches location may exhibit some artifacts.
\label{fig:results_mae}}
\end{figure}

\begin{table}[h]
\centering
\centering
\begin{tabular}{l|ccc}
\toprule
Method & Acc $\uparrow$ & Precision $\uparrow$  & F1 $\uparrow$  \\
\midrule
ViT\cite{dosovitskiy2021image} & 84.28$\pm$0.41\% &  84.17$\pm$ 0.44 \% & 84.25$\pm$ 0.42 \%  \\ 
ViT\cite{dosovitskiy2021image} + LC &   \underline{85.11$\pm$ 0.33\%}  & \underline{85.03$\pm$ 0.38\%}  & \underline{85.09$\pm$ 0.34\%}  \\
ViT\cite{dosovitskiy2021image} + LP + Reg & \textbf{85.35$\pm$ 0.35\%} &\textbf{85.33$\pm$ 0.37\%}  &  \textbf{85.34 $\pm$ 0.35\%} \\
\midrule
DWSNet\cite{navon2023equivariant} &  38.12$\pm$1.32\% &  36.33$\pm$1.54\%  &  37.11 $\pm$1.33\%  \\

Hyper \cite{schürholt2022hyperrepresentations} & 63.14 $\pm$ 1.12\% & 61.22 $\pm$ 1.45\%  & 62.45 $\pm$ 1.34\% \\

\bottomrule
\end{tabular}
\vspace{1mm}
\caption{\textbf{Refined CIFAR-10-INRs Classification.} We conduct additional experiments using refined CIFAR with 500 epochs.We report results from the weight-space-only method and observe that a performance gap still exists between the weights-only method and the RGB-based image method.
}
\label{tab:cifar_add_results}
\end{table}

\subsection{Additional information for Checklist}\label{sec_checklist}

\textbf{Potential negative societal impacts} While our work on proposing a large-scale INRs dataset for 2D and 3D tasks offers significant advancements in the field of implicit neural representations, it is important to consider potential negative societal impacts such like (1) Privacy Concerns: The proposed dataset use other popular public dataset and share the same risk for privacy violations of other dataset. (2) Our work limit only to natural images and more diverse modality should be considered. (3) Environmental Impact: Training large-scale INRs models requires significant computational resources, which can contribute to high energy consumption and increased carbon footprint. This environmental impact is a growing concern with the proliferation of large-scale AI models.

\small
\bibliographystyle{unsrt}
\bibliography{biblio}

\end{document}